\documentclass[manuscript]{acmart}
\AtBeginDocument{%
  \providecommand\BibTeX{{%
    \normalfont B\kern-0.5em{\scshape i\kern-0.25em b}\kern-0.8em\TeX}}}

\setcopyright{none}
\renewcommand\footnotetextcopyrightpermission[1]{}
\usepackage{booktabs}
\newcommand{\tabitem}{~~\llap{\textbullet}~~}
\usepackage{array}

\usepackage{makecell}

\usepackage{xcolor}
\usepackage{subcaption}

\begin{document}

\title{Human-Centered Explainable AI (XAI): From Algorithms to User Experiences}

 \author{Q. Vera Liao}
 \authornote{Reviewed work by the first author was done while working at IBM Research}
 \affiliation{%
   \institution{Microsoft Research}
   \city{Montreal}
   \country{Canada}}
 \email{veraliao@microsoft.com}

 \author{Kush R. Varshney}
 \affiliation{%
   \institution{IBM Research}
   \city{Yorktown Heights}
   \country{United States}}
 \email{krvarshn@us.ibm.com}
 
\renewcommand{\shortauthors}{Liao and Varshney}

\begin{abstract}
{\color{red}(Book Chapter Draft 4/2022) } In recent years, the field of explainable AI (XAI) has produced a vast collection of algorithms, providing a useful toolbox for researchers and practitioners to build XAI applications. With the rich application opportunities, explainability is believed to have moved beyond a demand by data scientists or researchers to comprehend the models they develop, to an essential requirement for people to trust and adopt AI deployed in numerous domains. However, explainability is an inherently human-centric property and the field is starting to embrace human-centered approaches. Human-computer interaction (HCI) research and user experience (UX) design in this area are becoming increasingly important. In this chapter, we begin with a high-level overview of the technical landscape of XAI algorithms, then selectively survey our own and other recent HCI works that take human-centered approaches to design, evaluate, and provide conceptual and methodological tools for XAI. We ask the question ``\textit{what are human-centered approaches doing for XAI}'' and highlight three roles that they play in shaping XAI technologies by helping navigate, assess and expand the XAI toolbox: to drive technical choices by users’ explainability needs, to uncover pitfalls of existing XAI methods and inform new methods, and to provide conceptual frameworks for human-compatible XAI. 

\end{abstract}

\maketitle

\section{Introduction}

In everyday life, people seek explanations when there is a gap of understanding. Explanations are sought for many goals that this understanding is meant to serve, such as predicting future events, diagnosing problems, resolving cognitive dissonance, assigning blame, and rationalizing one’s action. In interactions with computing technologies, an appropriate understanding of how the system works, often referred to as users' ``mental model''~\cite{norman2013design}, is the foundation for users to correctly anticipate system behaviors and interact effectively. A user’s understanding is constantly being shaped by what they see and experience with the system, and can be refined by being directly explained how the system works.

With the increasing adoption of AI technologies, especially popular inscrutable, opaque machine learning (ML) models such as neural networks models, understanding becomes increasingly difficult. Meanwhile, the need for stakeholders to understand AI is heightened by the uncertain nature of ML systems and the hazardous consequences they can possibly cause, as AI is now frequently deployed in high-stakes domains such as healthcare, finance, transportation, and even criminal justice. Some are concerned that this challenge of understanding will become the bottleneck for people to trust and adopt AI technologies. Others have warned that a lack of human scrutiny will inevitably lead to failures in usability, reliability, safety, fairness, and other moral crises of AI. 

It is with this overwhelming challenge of modern AI that the term explainable AI (XAI) and related terms such as AI ``interpretability'' and ''transpsrency'' have made their way into numerous academic works, industry efforts, as well as public policy and regulatory requirements. For example, the European Union General Data Protection Regulation (GDPR) now requires that ``meaningful information about the logic involved'' must be provided to people who are affected by automated decision-making systems. However, despite a vast collection of XAI algorithms produced by the AI research community and recent emergence of off-the-shelf toolkits (e.g.~\cite{AIX,arya2020ai,H2o,Microsoft,Skater}) for AI developers to incorporate state-of-the-art XAI techniques in their own models, successful examples of XAI are still relatively scarce in real-world AI applications.

Developing XAI applications is challenging because explainability, or the effectiveness of explanation, is not intrinsic to the model but lies in the perception and reception of the person receiving the explanation. Making a model completely transparent to its nuts and bolts does not guarantee that the person at the receiving end can make sense of all the information, or is not overwhelmed. What makes an explanation good---to provide appropriate information that can be understood and utilized---is contingent on the receiver’s current knowledge and their goal for receiving the explanation, among other human factors.

Therefore, developing XAI applications requires human-centered approaches that center the technical development on people’s explainability needs, and define success by human experience, empowerment, and well-being. It also means that XAI presents as much of a design challenge as an algorithmic challenge. Hence there are rich opportunities for HCI researchers and design practitioners to contribute insights, solutions, and methods to make AI more explainable. A research community of human-centered XAI~\cite{ehsan2021operationalizing,ehsan2020human,wang2019designing} has emerged, which bring in cognitive, sociotechnical, design perspectives, and more. We hope this chapter serves as a call to engage in this interdisciplinary endeavor by presenting a selected overview of recent AI and HCI works on the topic of XAI. While the growing collection of XAI algorithms offers a rich toolbox for researchers and practitioners to build XAI applications, we highlight three ways that human-centered approaches can help navigate, assess and expand this toolbox:

\begin{itemize}
    \item There is no one-fits-all solution in the growing collection of XAI techniques. The technical choices should be driven by users’ explainability needs, for which HCI and user research can offer methodological tools and insights about the design space (Section~\ref{needs}). 
    \item Empirical studies with real users can reveal pitfalls of existing XAI methods. To overcome the pitfalls requires both design efforts to fill the gaps and reflectively challenging fundamental assumptions in techno-centric approaches to XAI (Section~\ref{gaps}).
    \item Theories of human cognition and behaviors can offer conceptual tools to inspire new computational and design frameworks for XAI. However, this is still a nascent area and relevant theories in social science, behavioral science, and information science are yet to be explored (Section~\ref{theory}).
\end{itemize}{}

Before getting to these points, we will start with a brief overview of the technical landscape of XAI to ground our discussions (Section~\ref{techniques}). For interested readers, we suggest several recent papers that provide deeper technical surveys~\cite{guidotti2019survey,adadi2018peeking,arrieta2020explainable,carvalho2019machine}.

\section{What is explainable AI and what are the techniques?}
\label{techniques}
The definitions of explainability and related terms such as transparency, interpretability, intelligibility, and comprehensibility, are in a bit of flux. Scholars sometimes disagree on their scopes and how these terminologies intersect. However, XAI work often shares a common goal of \textit{making AI understandable by people}. By adopting this pragmatic, human-centered definition in the chapter, we consider XAI as broadly encompassing all technical means to this end of understanding, including direct interpretability, generating an explanation or justification, providing transparency information, etc. (and avoid the philosophical question ``what is or is not an explanation'' altogether~\cite{paez2019pragmatic}).  We note a distinction between a narrow scope of XAI focusing on explaining the model processes or internals, versus a broad scope that covers all explanatory information about the model, also including information about the training data, performance, uncertainty, and so on~\cite{liao2020questioning,vaughan2020human}. Since our focus is on XAI applications, we believe a broad view is necessary as users are often interested in a holistic understanding of the system and its behaviors. However, technical challenges are commonly presented in the inscrutability of the model internals, so the XAI techniques we review in this section are within the narrower scope of XAI.

It is worth mentioning that while the majority of XAI work focuses on ML models, and so does this chapter, there are emerging areas of other types of XAI including explainable planning, multi-agent systems, robotics, etc. In fact, the term XAI was coined half a century ago in the context of expert systems. We are currently in its second wave spurred by the popularity of ML technologies.

At a high level, XAI techniques fall into two camps~\cite{lipton2018mythos,guidotti2019survey}: 1) choosing a directly interpretable model, such as simpler models like decision trees, rule-based models, and linear regression; 2) choosing a complex, opaque model (sometimes referred to as ``black-box model''), such as deep neural networks and large tree ensembles, and then using a post-hoc technique to generate explanations. The choice between the two is sometimes discussed under the term ``performance-interpretability tradeoff'', as opaque models tend to perform better in many tasks. However, this tradeoff is not always true. Research has shown that in many contexts, especially with well-structured datasets and meaningful features, directly interpretable models can reach comparable performance to opaque models~\cite{rudin2019stop}. Moreover, an active research area of XAI focuses on developing new algorithms that possess both performance advantages and interpretable properties. For example, decision sets~\cite{lakkaraju2016interpretable}, generalized linear rule models~\cite{wei2019generalized}, GA2Ms~\cite{caruana2015intelligible}, and CoFrNets \cite{puri2021cofrnets} are recent algorithms that have more advanced computational properties than simple rule-based or linear models, but the model behaviors are still represented in meaningful rules or coefficients that can be understood relatively easily.  

However, opaque models are often chosen in practice because of their performance advantage for a given dataset, often a lower requirement for human effort (e.g., feature engineering), or the availability of off-the-shelf solutions. In these cases, one will have to use post-hoc XAI techniques to make the models explainable. Based on the purposes of explaining, \citet{guidotti2019survey} categorize post-hoc XAI techniques into \textit{global model explanations} on the overall logic of the model, \textit{outcome explanations} focusing on explaining a particular model output, and \textit{counterfactual inspection} that supports understanding how the model would behave with an alternative input\footnote{Note some adopt the categorization of \textit{global} versus \textit{local} explanations, and consider both outcome explanations and counterfactual explanations as local (concerned about the model decision process in a local region). }. Within these categories, XAI techniques commonly generate either \textit{feature-based} explanations to elucidate the model internals, or \textit{example-based} explanations to support case-based reasoning.

Note the three categories also apply to directly interpretable models. For example, a shallow decision-tree can be presented directly as a global explanation, highlighted of a particular path to explain a prediction outcome, or traced in alternative paths to perform counterfactual inspection. It is, however, much less straightforward with opaque-box models, which require separate post-hoc techniques, as we will give some examples below.

\textit{Examples of global model explanations.} Since it is impossible to understand the complex internals of an opaque model, the goal of a global model explanation is to provide an approximate overview of how the model behaves. This is often done by training a simple directly interpretable model such as a decision tree, rule set, or regression with the same training data, and performing optimization to make the simple model behave more closely to the original model.  For example, a technique called distillation changes the learning objective of the interpretable model to match the original model’s predictions~\cite{tan2018learning}.  Depending on the choice of the approximate model, global explanations can take the form of a decision tree, a set of rules, or feature weights. Without burdening people with the model decision complexities, showing representative examples and their predicted outcomes~\cite{yeh2018representer,papernot2018deep}, which overall have a good coverage representing the input space (whether constructed or extracted from the training data), can be used as example-based global explanations to shed light on how the model behaves for different kinds of input data.

\textit{Examples of outcome explanations.} To explain a prediction outcome made on an instance, a number of algorithms can be used to estimate the importance of each feature of this instance contributing to the prediction. For example, LIME (local interpretable model-agnostic explanations)~\cite{ribeiro2016should} starts by adding a small amount of noise to the instance to create a set of neighbor instances; it fits a simple linear model on those neighbors that mimics the original model’s behavior in the local region. The linear model’s weights can then be used as the feature importance to explain the prediction.  Another popular algorithm SHAP (Shapley additive explanations)~\cite{lundberg2017unified} defines feature importance based on Shapley values, inspired by cooperative game theory, to assign credit to each feature. Feature-importance explanations can be shown to users by visualizing the importance, or simply describing the most important features for the prediction. To explain deep neural networks, many other algorithms can be used to identify important parts of input features based on gradient~\cite{selvaraju2017grad}, propagation~\cite{bach2015pixel}, occlusion~\cite{li2016understanding}, etc. They are sometimes referred to as saliency methods and, when applied to image data, generate saliency maps. Example-based methods are useful to explain an outcome as well. For example, with some notion of similarity, finding similar instances in the training data with the same predicted outcome can be used to justify the prediction~\cite{kim2016examples,gurumoorthy2019efficient}.

\textit{Examples of counterfactual inspection.} Different from outcome explanations that describe the model's decision process for a given instance, counterfactual explanations--``counter to the facts''--are sought when people are interested in how the model would behave when the current input changes. In other words, people are interested in the ``why not a different prediction'' or ``how to change to get a different prediction''  questions rather than a descriptive ``why'' question. Such explanations are especially sought when seeking remedy or recourse for a current, often undesirable, outcome, such as ways to improve a patient's predicted high risk of a disease. Several algorithms can be used to generate conterfactual explanations by identifying changes, often with some notion of minimum changes, needed for an instance to receive a different prediction~\cite{dhurandhar2018explanations,looveren2021interpretable}. They are sometimes referred to as contrastive explanations for a counterfactual outcome (differentiated from counterfactual causes). Example-based methods can also be used to generate counterfactual examples--instances with minimum differences from the original one but having a different outcome~\cite{mothilal2019explaining,wachter2017counterfactual}. In other situations, people may want to zoom in on a specific feature and explore how its changes impact the model's prediction, i.e. asking a ``what if'' question. For this purpose, feature inspection techniques such as partial dependence plot (PDP)~\cite{hastie2009elements} and individual conditional expectation (ICE) can be used~\cite{goldstein2015peeking}.  

\vspace{5mm}
Most post-hoc techniques make some kind of approximation. Distillation and LIME approximate the complex model’s behaviors with a simpler model’s. PDP leaves out interactions between features. Example-based methods often explain by samples in the data. There is a long-standing debate regarding the potential risk of using approximate post-hoc techniques to explain instead of a directly interpretable model, as approximations will inevitably leave out some corner cases or even be unfaithful to what the original model computes~\cite{rudin2019stop}. 

However, in addition to the practical reasons mentioned earlier to opt for opaque-box models, there is a pragmatic argument to be made about the diverse communication devices people use to reach ``sufficient understanding'' to achieve a given objective. For example, to make a precise diagnosis of a problem, one may need explanations that describe a causal chain; whereas if one’s goal is to predict future events, following approximate rules or case-based reasoning could be sufficient and less cognitively demanding. One can also argue that when the model and the person have different epistemic access, approximations can be seen as a form of translation necessary to bridge the two. There is an emerging area of XAI research on generating human-consumable explanations with the supervision of human explanations~\cite{hind2019ted,ehsan2019automated,kim2018interpretability}, which essentially translates model reasoning into meaningful human explanations applied to the same prediction. This kind of explanation is a complete approximation but could be useful for laypeople who have difficulty understanding how ML models work, but want to get a sense of the validity of its predictions. We further highlight this objective and recipient dependent nature for the choice of explanation methods in the next sections.

That being said, developers of AI have a responsibility to understand, mitigate, and transparently communicate the limitations of approximate explanations to stakeholders. For example, an explainability metric known as faithfulness can be used to detect faulty post-hoc explanations~\cite{alvarez2018towards}. We must acknowledge that this is an actively researched topic and there is still a lack of principled approaches to identify and communicate the limitations of post-hoc explanations.

\section{Diverse explainability needs of AI stakeholders}
\label{needs}
It is easy to see that there are no ``one-fits-all'' solutions from this vast, and still rapidly growing, collection of XAI algorithms, and the choice should be based on target users’ explainability needs. The challenges are twofold: First, users of XAI are far from a uniform group and their explainability needs can vary significantly depending on their goals, backgrounds, and usage contexts. Second, XAI algorithms were often not developed with specific usage contexts in mind, or were developed primarily to help model developers or AI researchers inspect the models~\cite{miller2017explainable}. Hence their appropriateness to support end users’ explainability needs can be unclear. 

A starting point to address these challenges is to map out the design space of XAI applications and develop frameworks that account for people’s diverse explainability needs.  At a conceptual level, many attempt to summarize common user groups that demand explainability and what they would use AI explanations for~\cite{arrieta2020explainable,preece2018stakeholders,hind2019explaining}: 

\begin{itemize}
    \item \textbf{Model developers}, to improve or debug the model.
    \item \textbf{Business owners or administrators}, to assess an AI application’s capability, regulatory compliance, etc. to determine its adoption and usage.
    \item \textbf{Decision-makers}, who are direct users of AI decision support applications, to form appropriate trust in the AI and make informed decisions.
    \item \textbf{Impacted groups}, whose life could be impacted by the AI, to seek recourse or contest the AI.
    \item \textbf{Regulatory bodies}, to audit for legal or ethical compliance such as fairness, safety, privacy, etc.
    
\end{itemize}{}

While useful for considering different personas interacting with XAI systems, this kind of categorization lacks granularity to characterize people’s explainability needs. For example, a doctor using a patient risk-assessment AI (i.e., a decision-maker) would want to have an overview of the system during the onboarding stage, but delve into AI’s reasoning for a particular patient’s risk assessment when they treat the patient. Also, people in any of these groups may want to assess model capabilities or biases at certain usage points. 

In a recent HCI paper, \citet{suresh2021beyond} define a stakeholder's \textit{knowledge} and their \textit{objectives} as two components that cut across to determine one’s explainability needs. The authors characterize stakeholders’ knowledge by formal, instrumental, and personal knowledge and how it manifests in the contexts of machine learning, data domain, and general milieu. For stakeholders' goals and objectives, the authors propose a multi-level typology, ranging from long-term goals (building trust and understanding the model), immediate objectives (debug and improve the model, ensure regulatory compliance, take actions based on model output, justify actions influenced by a model, understand data usage, learn about a domain, contest model decisions), and specific tasks to perform with the explanations (assess the reliability of a prediction, detect mistakes, understand information used by the model, understand feature influence, understand model strengths and limitations).

While these efforts can be seen as top-down approaches to characterize the overall space of users' explainability needs, a complementary approach is to follow user-centered design and start with user research to identify application or interaction specific explainability needs. For example, \citet{eiband2018bringing} proposed a participatory design method that starts with analyzing users’ current mental model and gaps with how the system should be understood, based on an appropriate mental model prescribed by experts, to identify what needs to be explained.

In our own research with collaborators, we proposed to identify users' explainability needs by eliciting user questions to understand the AI~\cite{liao2020questioning}. This notion is based on prior HCI work using prototypical questions to represent ``intelligibility types''~\cite{lim2009assessing}, and social science literature showing that people’s explanatory goals can be expressed in different kinds of questions~\cite{hilton1990conversational}. By interviewing 20 designers, we collected common questions users ask across 16 ML applications and developed an \textit{XAI Question Bank}, with more than 50 detailed user questions organized in 9 categories: 

\begin{itemize}
    \item \textbf{How} (global model-wide): asking about the general logic or process the AI follows to have a global view.
    \item \textbf{Why} (a given prediction): asking about the reason behind a specific prediction.
    \item \textbf{Why Not} (a different prediction): asking why the prediction is different from an expected or desired outcome.
    \item \textbf{How to be That} (a different prediction)\footnote{The difference between \textit{Why Not} and \textit{How to Be That} can be subtle and context-dependent. Users may ask \textit{Why Not} when seeing an unexpected prediction and interested in comparing what gets the counterfactual outcome. Users may ask \textit{How to Be That} when seeking recourse so the explanation should more specifically focus on minimum or actionable changes they can make to the current input.}: asking about ways to change the instance to get a different prediction.
    \item \textbf{How to Still Be This} (the current prediction): asking what change is allowed for the instance to still get the same prediction.
    \item \textbf{What if}: asking how the prediction changes if the input changes.
    \item \textbf{Performance}: asking about the performance of the AI.
    \item \textbf{Data}: asking about the training data.
    \item \textbf{Output}: asking what can be expected or done with the AI’s output.

\end{itemize}{}

These questions once again demonstrate that XAI should be defined broadly, not limited to explaining model internals, as users are also interested in explanatory information about the performance, data, and scope of output, among other dimensions.

This XAI question bank maps out the space of common explainability needs and can be used as a tool to identify applicable questions in user research. In a follow-up work~\cite{liao2021question}, we propose a \textit{question-driven user centered design} method that starts with identifying key user questions by user research, then uses these questions to guide the choices of XAI techniques and iterative design. To facilitate this process and foreground users' explanability needs, we suggest to reframe the technical space of XAI by the user question that each XAI technique can address. For example, a feature-importance explanation technique can answer the \textit{Why} question, while a counterfactual explanation can answer the \textit{How to be That} question. We provide a suggested mapping between the question categories and example XAI techniques, as shown in Table~\ref{tab:mapping}, focusing on techniques that are available in current open-source XAI toolkits accessible for practitioners~\cite{AIX,Skater,H2o,Microsoft}. We suggest that UX practitioners and data scientists should engage in collaborative discussions to identify appropriate XAI techniques, using the mapping guidance as a reference. Based on these technical choices and insights from users questions, XAI user experience can be designed and evaluated in an iterative fashion.

\vspace{5mm}
In short, a growing collection of XAI techniques offer a rich toolbox for researchers and practitioners to build XAI applications. Making effective and responsible choices from this toolbox should be guided by users’ explainability needs. HCI research offers means to understand user needs for specific applications, insights about real-world user needs to better frame and organize this toolbox, as well as methodological tools to help navigate the toolbox. In the next section, we discuss how HCI research can also inform the limitations of the current technical XAI toolbox.

\begin{table}[]
    \centering
    {\small
    \begin{tabular}{>{\centering\arraybackslash}m{2.6cm}|m{7.9cm}|m{3.6cm}}
    \toprule
        \multicolumn{1}{>{\centering\arraybackslash}m{2.8cm}|}{Question} &\multicolumn{1}{>{\centering\arraybackslash}m{8cm}|}{Ways to explain} & \multicolumn{1}{>{\centering\arraybackslash}m{3cm}}{Example XAI methods}\\
        
         \midrule
         
       \makecell{\textbf{How}\\(global model-wide)}&\tabitem Describe the general model logic as feature impact$^\ast$, rules$^\dagger$ or decision-trees$^\ddagger$  \newline\tabitem If user is only interested in a high-level view, describe what are the top features or rules considered &  ProfWeight$^{\ast\dagger\ddagger}$~\cite{dhurandhar2018improving}, Global feature importance$^\ast$~\cite{wei2015variable,lundberg2020local}, Global feature inspection plots$^\ast$ (e.g. PDP~\cite{hastie2009elements}), Tree surrogates$^\ddagger$~\cite{craven1995extracting}
       \\
        \midrule
        
       \makecell{\textbf{Why}\\(a given prediction)} &\tabitem Describe how features of the instance, or what key features, determine the model's prediction of it$^\ast$  \newline\tabitem Or describe rules that the instance fits to guarantee the prediction$^\dagger$ \newline\tabitem Or show similar examples with the same predicted outcome to justify the model’s prediction$^\ddagger$ &
       LIME$^\ast$ \cite{ribeiro2016should},  SHAP$^\ast$ \cite{lundberg2017unified},  LOCO$^\ast$ \cite{lei2018distribution},  Anchors$^\dagger$ \cite{ribeiro2018anchors},  ProtoDash$^\ddagger$ \cite{gurumoorthy2019efficient}
       \\
       \midrule

       \makecell{\textbf{Why Not}\\(a different prediction)}&\tabitem Describe what features of the instance determine the current prediction and/or with what changes the instance would get the alternative prediction$^\ast$ \newline\tabitem Or show prototypical examples that have the alternative outcome$^\dagger$ & CEM$^\ast$ \cite{dhurandhar2018explanations},  Counterfactuals$^\ast$ \cite{looveren2021interpretable},  ProtoDash$^\dagger$ (on alternative prediction) \cite{gurumoorthy2019efficient}
       \\
    \midrule

       \makecell{\textbf{How to Be That}\\(a different prediction)}&\tabitem Highlight feature(s) that if changed (increased, decreased, absent, or present) could alter the prediction to the alternative outcome, with minimum effort required$^\ast$  \newline\tabitem Or show examples with minimum differences but had the alternative outcome$^\dagger$& CEM$^\ast$ \cite{dhurandhar2018explanations},  Counterfactuals$^\ast$ \cite{looveren2021interpretable},  Counterfactual instances$^\dagger$ \cite{wachter2017counterfactual},  DiCE$^\dagger$ \cite{mothilal2019explaining}
       \\
    \midrule

       \makecell{\textbf{How to Still Be This}\\(the current prediction)}&\tabitem Describe features/feature ranges$^\ast$  or rules$^\dagger$ that could guarantee the same prediction \newline\tabitem Or show examples that are different from the instance but still had the same outcome&CEM$^\ast$~\cite{dhurandhar2018explanations}, Anchors$^\dagger$~\cite{ribeiro2018anchors}
       \\
               \midrule
        
   \textbf{What if} &\tabitem Show how the prediction changes corresponding to the inquired change of input &PDP~\cite{hastie2009elements},  ALE~\cite{apley2020visualizing}, ICE~\cite{goldstein2015peeking}
       \\
    \midrule
        
       \textbf{\textit{Performance}} &\tabitem Provide performance information of the model \newline\tabitem Provide uncertainty information for each prediction$^\ast$ \newline\tabitem Describe potential strengths and limitations of the model &
       Precision, Recall, Accuracy, F1, AUC; Communicate uncertainty of each prediction$^\ast$~\cite{ghosh2021uncertainty}; See examples in FactSheets~\cite{arnold2019factsheets} and Model Cards~\cite{mitchell2019model}
\\
          \midrule
        
       \textbf{\textit{Data}} &\tabitem Provide comprehensive information about the training data, such as the source, provenance, type, size, coverage of population, potential biases, etc. & See examples in FactSheets~\cite{arnold2019factsheets} and Datasheets~\cite{gebru2021datasheets}
       \\
       
                 \midrule
        
       \textbf{\textit{Output}} & \tabitem Describe the scope of output or system functions.
\newline\tabitem If applicable, suggest how the output should be used for downstream tasks or user workflow & See examples in FactSheets~\cite{arnold2019factsheets} and Model Cards~\cite{mitchell2019model}
       \\
       
        \bottomrule
      
    \end{tabular}
    }
    \caption{A mapping guidance between categories of user questions in XAI question bank~\cite{liao2020questioning} and example XAI methods to answer these questions, with descriptions of their output in ``Ways to explain'' column. XAI methods are selected based on what are available in current open-source XAI toolkits~\cite{AIX,Microsoft,Skater,H2o}. The last three rows (in \textit{italic}) are broader XAI needs not limited to explaining model processes. This mapping guidance can support identifying appropriate XAI techniques based on user questions.  }
    \label{tab:mapping}
\end{table}

\section{Pitfalls of XAI: Minding the gaps between algorithmic explanations and actionable understanding}
\label{gaps}
With so many XAI algorithms developed, one must ask: do they work? The answer is complicated because of the diverse contexts where XAI is sought for. The answer is also difficult because it requires understanding how people perceive, process, and use AI explanations. HCI research, and more broadly human-subject studies, are key to evaluating XAI in the context of use~\cite{doshi2017towards}, identifying where it falls short, and informing human-centered solutions. While many studies showed positive results that XAI techniques can improve people’s understanding of the model~\cite{lucic2020does,ribeiro2018anchors,hase2020evaluating,lakkaraju2017interpretable}, in this section we draw attention to a few pitfalls of XAI based on recent HCI research, and highlight two sources of disconnect leading to the pitfalls.

\paragraph{The first source of XAI pitfalls is a disconnect between technical XAI approaches and supporting users' end goals in usage contexts,} as many empirical studies failed to find conclusive evidence that adding XAI components improve performance for realistic AI-assisted user tasks such as judgment and decision making~\cite{zhang2020effect,bansal2021does,poursabzi2021manipulating,wang2021explanations} and image and video analysis~\cite{fan2022human,ghassemi2021false}. This disconnect must be critically examined by taking the position that users’ goal with XAI is not an understanding defined in a vacuum, but an \textit{actionable understanding} that is sufficient to serve the objective that they seek explanations for. However, the diverse and dynamic user objectives that we discussed earlier are often not explicitly considered when developing XAI algorithms. Some even criticized the technical XAI field insofar as having an ``inmates running the asylum'' problem~\cite{miller2017explainable,miller2018explanation}, as AI researchers often develop algorithms based on their own intuition of what constitutes good explanations rather than the needs of intended users.

This disconnect is reflected in the current practices of how XAI algorithms are evaluated, which can profoundly
shape the output of the field. While progress has been made by recognizing the necessity of human evaluation beyond earlier focuses solely on algorithmic properties, a recent study by~\citet{buccinca2020proxy} points out that ``proxy tasks'' widely used by AI researchers to evaluate their proposed XAI techniques can be misleading. A common example of a proxy task is a ``simulation test'', which asks people to predict the model’s output based on the input and the explanation. Such tests, without specifying an end goal of understanding, can fail to predict the effectiveness of XAI techniques in real tasks that people seek explanations for, such as improving decision-making.

There can be many reasons for this divide between effectiveness in proxy tasks and deployment. As~\citet{buccinca2020proxy} point out, performing proxy tasks in a controlled setting could induce a different cognitive process from a realistic setting, such as granting more attention to the explanation. Moreover, the ability to simulate a model prediction may simply not match the need for a user to perform a realistic task. For instance, in the context of decision-making, the key to the success of a human-AI team is appropriate reliance, i.e.\ knowing when to trust the AI’s recommendation and when to be cautious. An actionable understanding for appropriate reliance requires not only knowing how the model makes predictions, but also how to judge if the reasoning is flawed, for which a user may lack either the knowledge or cognitive capacity. Filling this gap of understanding may require a different kind of transparency. For example, recent HCI studies repeatedly find that showing uncertainty information of individual predictions is more effective than local outcome explanations to help people form appropriate reliance on AI~\cite{zhang2020effect,bansal2021does}.

\begin{figure}
    \begin{minipage}{0.45\textwidth}
        \begin{subfigure}{\linewidth}
            \includegraphics[width=\textwidth]{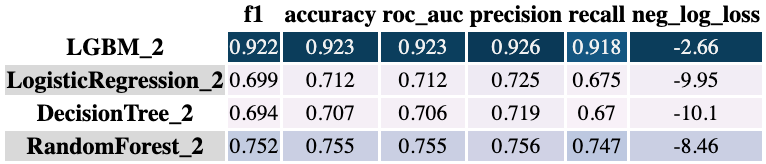}
            \caption{Screenshot of the Metrics Table showing metrics for four selected models.}
            \label{fig:metrics-table}
        \end{subfigure}
        \par\bigskip
        \begin{subfigure}{\linewidth}
            \includegraphics[width=\textwidth]{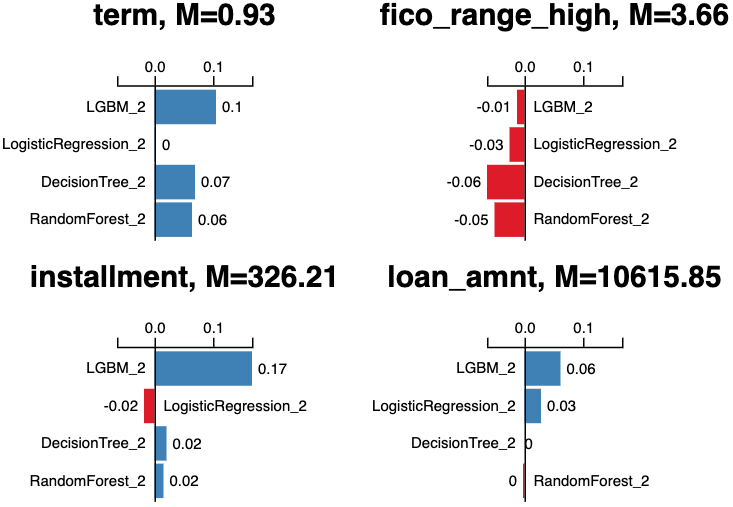}
            \caption{Partial screenshot of the Feature Importance Comparison View showing 4 of 21 FI plots.}
            \label{fig:lfc}
        \end{subfigure}
    \end{minipage}
    \hfill
    \begin{minipage}{0.5\textwidth}
    \begin{subfigure}{\linewidth}
        \includegraphics[width=\textwidth]{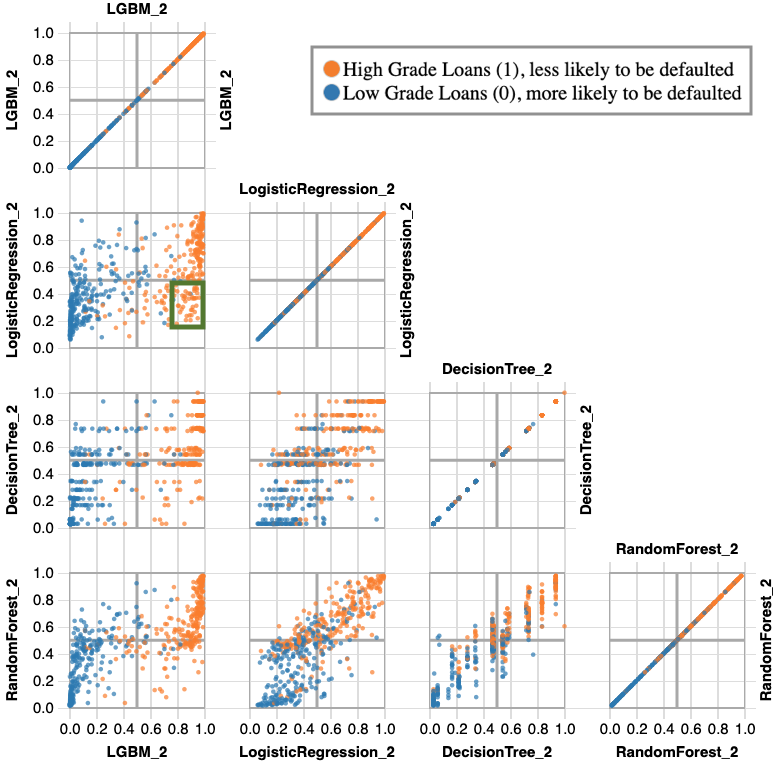}
        \caption{Screenshot of the Probability Scatterplot Matrix displaying pairwise comparisons of 4 models.}
        \label{fig:psm}
    \end{subfigure}
    \end{minipage}
    \caption{Model LineUpper, an example XAI tool in~\citet{narkar2021model} that supports ML developers to compare multiple candidate models by comparing their feature-importance explanations at multiple levels---at a global level, for a region of input space, or a specific instance, by selecting from the right-side Scatterplot Matrix panel.}
    \label{fig:lineUpper}
\end{figure}


To close this gap between technical XAI and user experiences requires both studying user interactions with XAI in the contexts of use, and operationalizing human-centered perspectives in XAI algorithms, including developing evaluation methods that better account for the actual user needs in different downstream usage contexts. 

\begin{figure*}
  \centering
  \includegraphics[width=0.8\columnwidth]{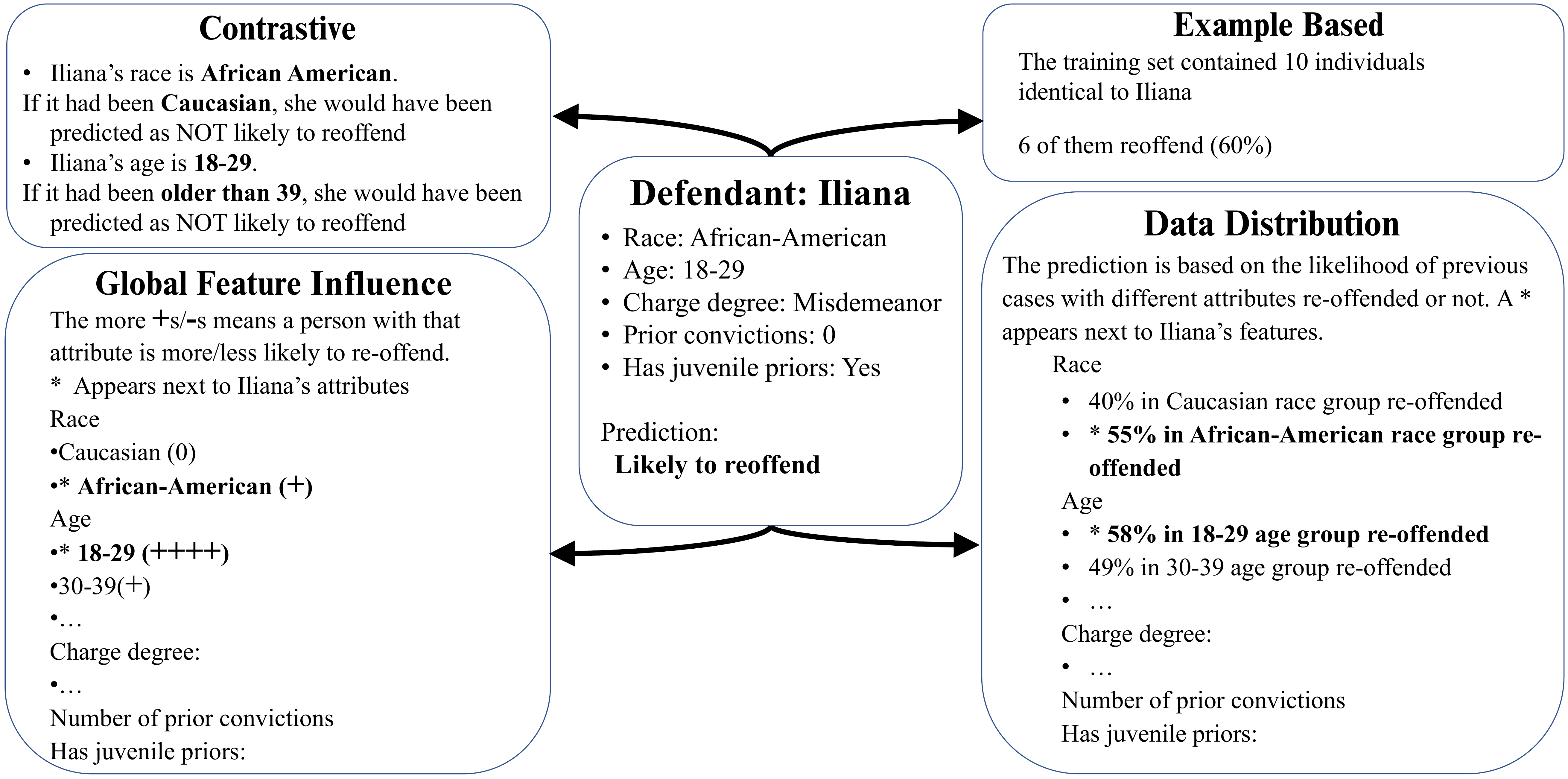}
   \vspace{-1em}
  \caption{Four types of XAI features compared in~\citet{dodge2019explaining} (with minor updates on the names of explanations from the original paper) to support people's fairness judgment of ML models, with a use case of an ML model performing recidivism risk predictions. Contrastive explanation (top left) focuses on how the defendant would need to change to be predicted to have low risk. It is more effective in revealing individual fairness issues of an unfair model--similar individuals from different protected groups are treated differently--than the two global explanations at the bottom. Example based explanation can suggest fairness issues by revealing the ambivalence of the decision (only 60\% of a similar profile re-offend).  }~\label{fig:fairness}
   \vspace{-2em}
\end{figure*}

For the first part, recent HCI research provides useful insights for supporting XAI user experiences in some common usage contexts. For model development or debugging, research suggests that users often need a range of explanations for different levels of the model behaviors to perform a comprehensive diagnosis~\cite{hohman2019gamut,narkar2021model,hong2020human}. For example, Figure~\ref{fig:lineUpper} shows \textit{Model LineUpper}~\cite{narkar2021model}, an XAI tool that we designed with our collaborators to support data scientists to compare multiple models (in the context of choosing from multiple candidate models generated by automated machine learning (AutoML)) by comparing their feature-importance explanations. This comparison can happen at different levels: for a global view, for an input region selected on the Scatterplot Matrix on the right, and down to individual instances. For decision-makers, such as users of an AI system supporting medical diagnosis~\cite{cai2019hello,xie2020chexplain,liao2020questioning} , they may want to see a high-level global model explanation during the onboarding stage to form an appropriate mental model, but have higher demand for outcome explanations and counterfactual inspection, especially when they get unexpected or suspicious predictions from the model. When it comes to auditing for model fairness, our work with collaborators compared the effectiveness of four types of explanation (shown in Figure~\ref{fig:fairness}) and found that contrastive explanations can effectively help people identify concerns of individual fairness, where similar individuals from different protected groups such as races are treated differently by the model~\cite{dodge2019explaining}.

\paragraph{The second source of XAI pitfalls brought to light by empirical research is a disconnect between assumptions underlying technical approaches to XAI and people's cognitive processes} A pitfall robustly found in recent work is that \textit{explanations can lead to unwarranted trust or confidence in the model}. In a controlled experiment where an ML model was used to assist participants to predict apartment sales prices, \citet{poursabzi2021manipulating} found that, contrary to the hypothesis, showing people an explainable model with feature importance hindered their ability to detect model mistakes. By conducting a contextual inquiry study with data scientists using popular XAI techniques (e.g. SHAP) during model development, \citet{kaur2020interpreting} found that the existence of explanations could mistakenly lead to over-confidence that the model is ready for deployment. In the context of a nutrition recommender, \citet{eiband2019impact} showed that even placebic explanations, which did not convey useful information, invoked a similar level of trust as real explanations did. In addition, there is the concern of illusory understanding, with which one subjectively over-estimates the understanding they gain from XAI~\cite{chromik2021think}. Explanations can also create information overload and distract people from forming a useful mental model of how a system operates~\cite{springer2019progressive}.

These observations highlight the danger of deploying technologies without a clear understanding of how people interact with them. One way to move the field forward is to connect with theories and insights about human behaviors and cognition. For example, dual-process theories~\cite{kahneman2011thinking,petty1986elaboration} provide a critical lens to understand how people process XAI. The central thesis of these theories is that people can engage in two different systems to process information and make decisions. System 1 is intuitive thinking, often following mental shortcuts and heuristics; System 2 is analytical thinking, relying on careful reasoning of information and arguments. Because System 2 is slower and more cognitively demanding, people often resort to System 1 thinking, which, when applied inappropriately, can lead to cognitive biases and sub-optimal decisions. Through this theoretical lens, there is an increasing awareness~\cite{buccinca2020proxy,wang2019designing,ehsan2021explainable,nourani2021anchoring,rastogi2022deciding}  that while XAI techniques make an implicit assumption that people can and will attend to every bit of explanations, in reality, people are more likely to engage in System 1 thinking. 

However, \textit{it remains an open question of what kind of heuristics can be triggered by XAI with System 1 thinking}. It is possible that people superficially associate the ability to provide explanations directly with competence, and therefore form unwarranted trust and confidence. Heuristics are developed through past experiences, and can evolve as people experience new technologies. \citet{nourani2021anchoring} demonstrate that when interacting with XAI, people are vulnerable to common cognitive biases such as anchoring bias after observing model behaviors early on. A recent study by~\citet{ehsan2021explainable} uncovered diverse heuristics people follow in response to AI explanations, such as associating explanations with affirmation and social presence, and associating a specific presentation of explanation--numerical numbers--with intelligence and algorithmic thinking. 

Another critical implication of dual-process theories is that people do not equally engage in System 1 or System 2 thinking in all contexts. People are inclined to engage in System 1 thinking when they lack either the ability or motivation (broadly defined) to perform analytical thinking~\cite{petty1986elaboration}. This difference can lead to \textit{another pitfall of XAI--potential inequalities of experience} including the risk of mistrust and misuse of AI. For example, a study found that AI novices, compared to experts, not only had less performance gain from XAI but were also more likely to have illusory satisfaction~\cite{szymanski2021visual}. Other studies suggest that in time and cognitive resource constrained settings, people are less able to process explanations effectively~\cite{xie2020chexplain,robertson2021wait}. In our own work with collaborators \cite{ghai2021explainable}, we showed that adding explanations in an active learning setting (i.e. label instances requested by the model) decreased satisfaction for people who scored low in Need for Cognition, a personality trait reflecting one’s general motivation to engage in effortful thinking.

Research has begun to address this mismatch between people’s cognitive processes and assumptions underlying XAI. One way is to provide interventions to nudge people to engage deeper in System 2 thinking. \citet{buccinca2021trust} introduced cognitive forcing functions as design interventions for that purpose, including asking users to make decisions before seeing the AI’s recommendations, slowing down the process, and letting users choose when to see the AI recommendation. In our own work~\cite{rastogi2022deciding}, we also saw that increasing the time for users to interact with the ML system mitigated System 1 biases. Another path is to seek technical and design solutions that reduce the cognitive workload imposed by XAI, by reducing the quantity and improving the consumability of information. For example, studies suggest that muti-modalities (text, visual, audio, etc.) can be leveraged to aid attention and understanding of XAI~\cite{robertson2021wait,szymanski2021visual}. Progressive disclosure~\cite{springer2019progressive}, starting with simplified or high-level transparency information and revealing details later or upon user requests, is another effective approach to reduce cognitive workload. Technical approaches that optimize for a balance between explanation accuracy and conciseness have also been explored~\cite{abdul2020cogam}.

We must note that heuristics are an indispensable part of people’s decision-making process. If applied appropriately, they can aid people to make more efficient and optimal decisions. In fact, they may be key to closing the inequality gaps for people without an ideal profile of ability or motivation to process information about AI. For example, we may envision a quality endorsement feature through some authorized third-party inspecting a model with explanations. This could allow laypeople to apply a reliable ``authority heuristic'' and defer to the experts' judgments. Understanding what heuristics are involved in interactions with XAI and AI in general, and how to leverage reliable heuristics to improve human-AI interaction, are important open questions for the field.

\vspace{5mm}
We close this section with an optimistic note that by centering our analysis on people, on how they interact with and process information about AI, and whether they can achieve their objectives, we can move away from a techno-centric focus on generating algorithmic explanations. We can begin to identify opportunities to improve user experiences in this currently under-developed space between algorithmic explanations and actionable understanding, and appreciate explainable AI as much of a design problem as a technical problem. The design solutions may be concerned with how to communicate algorithmic explanations effectively, such as choosing the right modalities, level of abstraction, workarounds for privacy or security constraints, and so on. They may also come in the form of interventions to influence how people process XAI, such as providing cognitive forcing functions or guidance that help people better assess explanatory information~\cite{rieh2007credibility}. Furthermore, it is necessary to fill the knowledge or information gaps for users to achieve actionable understanding beyond algorithmic explanations, such as providing necessary domain knowledge (e.g. what a feature means) and general notions of how AI works.

\begin{figure*}
  \centering
  \includegraphics[width=0.8\columnwidth]{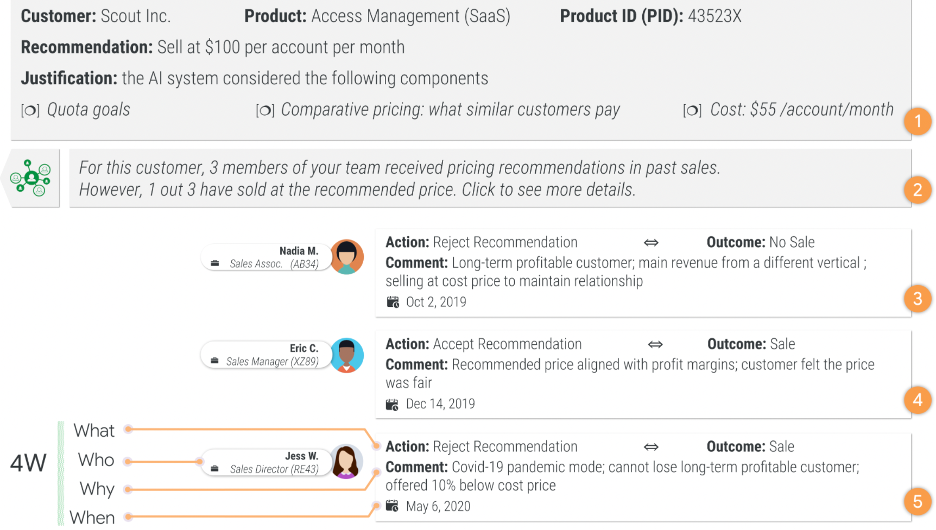}
   \vspace{-1em}
  \caption{A scenario-based design of Social Transparency in AI systems as used in~\cite{ehsan2021expanding}. It combines technical explanations and ``4W features'' (What, Who, Why. and When) that reflect the historical decision trajectory of other users to provide transparency into the sociotechnical system. }~\label{fig:ST}
   \vspace{-2em}
\end{figure*}

While we discussed the pitfalls of XAI mostly through a cognitive lens, implicit in supporting actionable understanding is a requirement to approach XAI as a sociotechnical problem~\cite{ehsan2020human}, especially given that consequential AI systems are often embedded in socio-organizational contexts with their own history, shared knowledge and norms. On the one hand, for XAI technology developers, to understand the ``\textit{who}'' in XAI and articulate their needs and objectives requires situating the ``\textit{who}'' in the sociotechnical context. On the other hand, for XAI users, an actionable understanding is often a \textit{socially situated understanding}, which enables them to make sense of not only the technical component but also the sociotechnical system as a whole. Motivated by this sociotechnical perspective, with collaborators we proposed the concept of social transparency--making visible the social-organizational factors that govern the use of AI systems~\cite{ehsan2021expanding}. Operationalized in a design framework to present past users’ interactions and reasoning with the AI (see a design in Figure~\ref{fig:ST} studied in~\cite{ehsan2021expanding}), we demonstrated that such information could help users make more informed decisions and improve the collective experience with AI as a sociotechnical system.

\section{Theory-Driven Human-Compatible XAI}
\label{theory}
Previously we gave an example of using dual-process theories to retrospectively understand how people interact with XAI. In this section we discuss another important human-centered approach to XAI, by performing theoretical analysis of human explanations, as well as broader cognitive and behavioral processes, to inspire new computational and design frameworks to make XAI more human-compatible. 

Such work is best represented by Miller’s seminal paper that brings insights from social sciences about fundamental properties of human explanations to the common awareness of the AI community~\cite{miller2018explanation}. By surveying a large volume of prior work on how people seek, generate, and evaluate explanations in philosophy, psychology, and cognitive science, Miller summarized four primary properties of human explanations: 1) Explanations are often contrastive, sought in response to some counterfactual cases. This is because a  \textit{Why} question is often triggered by ``abnormal or unexpected'' events, asked to understand the cause of an event relative to some other event that did not occur. In other words, the \textit{Why} question is often an implicit \textit{Why not} question. 2) Explanations are selected by the explainer, often in a biased manner. When explaining to others, people rarely give an actual or complete cause of an event, but select a small number of causes based on some criteria or heuristics. 3) Explanations are social, as a transfer of knowledge, often part of a conversation or interaction, and presented relative to the explainer’s beliefs about the explainee’s beliefs. 4) Using probabilities or statistical information to explain is often ineffective and unsatisfying. Explicitly referring to causes is often more effective.

Published in 2019, in just two years, this work has made a significant impact on the XAI field. For instance, the point about explanations being contrastive has inspired many to work on counterfactual explanations to answer the \textit{Why Not} and \textit{How to be That} questions, as we reviewed in Section~\ref{techniques}. A recent work by~\citet{alvarez2021human} attempts to operationalize contrastiveness and selectiveness by a ``weight of evidence'' framework adapted from information theory. This technique allows users to flexibly inspect current prediction against multiple alternative hypotheses they choose in a compositional way (e.g., ``a fever rules out a cold in favor
of bronchitis or pneumonia; among these, chills suggest the latter''), which is especially suitable for multi-class settings.

From a user-interaction point of view, the points of explanations being selected and social have profound implications. Miller reviewed several useful theories about how people present explanations to others, which we believe can provide conceptual grounds to frame XAI as an interaction problem. One of them is Malle’s theory of explanation~\cite{malle2006mind}, which breaks the generation of explanations into two distinct and co-influencing psychological processes: 1) Information process for the explainer (i.e. AI in the case of XAI) to devise explanations, which is determined by what kind of information the explainer has access to. 2) Impression management process that governs the social interactions with the explainee (i.e., users in the case of XAI), which is driven by the pragmatic goal of the explainer, such as transferring knowledge, generating trust in the explainee, assigning blame, and so on.

With this conceptualization, current technical XAI approaches are primarily concerned with the information process, leaving the impression management process an open area yet to be explored, which we believe is key to making XAI effectively selected, less cognitively demanding, and more consumable in general. A useful set of resources to inform XAI work on this topic, as Miller suggested, is to look into the processes of how people select explanations from available causes, by following common heuristics such as abnormality (selecting the abnormal cause), intentionality (select intentional actions), necessity, sufficiency, and robustness (selecting causes that would hold in many situations). The choice highly depends on the explainer's goal, which again highlights the importance of specifying the objective of explaining. Further, we point to broader social science research on impression management~\cite{goffman1978presentation,leary1990impression}, on influencing others' perception by regulating information in social interactions, as well as ethics discussions around it, to draw inspiration from.

The social nature of explanation also maps to an essential requirement for \textit{interactivity} in XAI applications \cite{krause2016interacting}. User interactions do not end at receiving an XAI output, but continue until an actionable understanding is achieved. In other words, as users’ explainability needs are expressed in questions, they will keep asking follow-up questions until satisfied, thus engaging in back-and-forth conversations. Conversational models of explanation, as well as general principles of conversations and communication (e.g., Grice’s maxims that a speaker follows to optimize for the desired social goal~\cite{grice1975logic}; theory of grounding in communication~\cite{clark1991grounding}), hold promise for informing technologies and design for interactive XAI. Miller reviewed several relevant theories including Hilton’s conversational model of explanations~\cite{hilton1990conversational}, which postulates that a good explanation must be relevant to the focus of a question and present a topology of different causal questions. \citet{antaki1992explaining} extended this model to a wider class of argumentative dialogue for the common pattern of claim-backing in explanations. \citet{walton2004new} further extended this line of work into a formal dialogue model of explanation, including a set of speech act rules. These theories offer appealing grounds to build computational models, and recent XAI has begun to explore dialogue models for interactive XAI~\cite{madumal2019grounded}. Outside XAI, work on dialogue systems frequently builds on formal models of human conversational and social interactions (e.g.~\cite{bickmore2001relational}), including systems that generate explanatory dialogues~\cite{cawsey1992explanation}. 

Theories can also inform design frameworks that guide researchers and practitioners to investigate the design space and make design choices. For example, \citet{wang2019designing} performed a comprehensive analysis on the theoretical underpinning of human reasoning to derive a conceptual framework that allows linking XAI methods to users’ reasoning needs. This framework includes four dimensions that describe a normative view of how people should reason with explanations, including explanation goals, reasoning process, causal explanation type, and elements in rational choice decisions. It also separately describes people’s natural decision-making, and the errors and limitations they are subject to, based on dual-process theories. Designers can use the framework to perform a conceptual analysis to understand, e.g., based on user research, users’ reasoning goals and potential errors, to identify what XAI methods can support their goals, or to investigate gaps in current XAI methods. The authors further provide a mapping between elements under these human-reasoning dimensions and existing XAI approaches, and guidelines on how to use XAI methods to mitigate common cognitive biases.

While hugely promising, theory-driven XAI is still a nascent area. Many areas of cognitive, social, and behavioral theories are yet to be explored. For example, given that users of XAI are information seekers to achieve actionable understanding, theories in information science such as models of sense-making~\cite{dervin1998sense} and information seeking behaviors~\cite{wilson1981user} (how people’s information needs drive their behaviors and information use) can offer useful theoretical lenses to formalize and anticipate user behaviors. However, the major challenge lies in how to operationalize theoretical insights and formal behavioral models into computational and design frameworks, which may require, as many have already argued~\cite{miller2018explanation,doshi2017towards,vaughan2020human}, collaboration across the research disciplines of AI, HCI, social sciences and more.  

\section{Summary}

Explainable AI is one of the fastest-growing areas of AI in several directions: a rapidly expanding collection of techniques, substantial industry efforts to produce open-source XAI toolkits for practitioners to use, and widespread public awareness and interest in the topic. It is also a fast-growing area for human-centered ML, which can be seen in a proliferation of XAI research published in HCI and social science venues in recent years. Adopting human-centered approaches to XAI is inevitable given that explainability is a human-centric property and XAI must be studied as an interaction problem. However, different from some other topics in this book, HCI work on XAI currently resides in, and often needs to challenge, a techno-centric reality given that the technical AI community has made strides already. A research community of human-centered XAI~\cite{ehsan2021operationalizing,ehsan2020human,wang2019designing} has emerged. In this chapter we provide a selected overview on works from this emerging community to help researchers and practitioners understand insights, available resources, and open problems in utilizing XAI techniques to build XAI user experiences. We hope this chapter could encourage future research to continue bridging design practices and state-of-the-art XAI techniques, uncovering pitfalls of and challenging algorithmic assumptions, and building human-compatible XAI from theoretical grounds. We also hope these approaches will inspire work to address broader challenges in human-centered ML. 

\begin{acks}
We thank Upol Ehsan, Tim Miller, Jenn Wortman Vaughan and Editors of the book for their generous feedback. We are also grateful to members of the Human-AI Collaboration group and the Foundations of Trustworthy AI department at IBM Research -- Thomas J. Watson Research Center, whose work and conversations with us shaped our thinking. 
\end{acks}
\bibliographystyle{ACM-Reference-Format}
\bibliography{refs}


\begin{thebibliography}{109}


\ifx \showCODEN    \undefined \def \showCODEN     #1{\unskip}     \fi
\ifx \showDOI      \undefined \def \showDOI       #1{#1}\fi
\ifx \showISBNx    \undefined \def \showISBNx     #1{\unskip}     \fi
\ifx \showISBNxiii \undefined \def \showISBNxiii  #1{\unskip}     \fi
\ifx \showISSN     \undefined \def \showISSN      #1{\unskip}     \fi
\ifx \showLCCN     \undefined \def \showLCCN      #1{\unskip}     \fi
\ifx \shownote     \undefined \def \shownote      #1{#1}          \fi
\ifx \showarticletitle \undefined \def \showarticletitle #1{#1}   \fi
\ifx \showURL      \undefined \def \showURL       {\relax}        \fi
\providecommand\bibfield[2]{#2}
\providecommand\bibinfo[2]{#2}
\providecommand\natexlab[1]{#1}
\providecommand\showeprint[2][]{arXiv:#2}

\bibitem[\protect\citeauthoryear{??}{H2o}{2017}]%
        {H2o}
 \bibinfo{year}{2017}\natexlab{}.
\newblock \bibinfo{title}{H2O.ai Machine Learning Interpretability}.
\newblock
\newblock
\newblock
\shownote{https://github.com/h2oai/mli-resources.}


\bibitem[\protect\citeauthoryear{??}{Ska}{2018}]%
        {Skater}
 \bibinfo{year}{2018}\natexlab{}.
\newblock \bibinfo{title}{Model Interpretation with Skater}.
\newblock
\newblock
\newblock
\shownote{https://oracle.github.io/Skater/.}


\bibitem[\protect\citeauthoryear{??}{AIX}{2019}]%
        {AIX}
 \bibinfo{year}{2019}\natexlab{}.
\newblock \bibinfo{title}{IBM AIX 360}.
\newblock
\newblock
\newblock
\shownote{aix360.mybluemix.net/.}


\bibitem[\protect\citeauthoryear{??}{Mic}{2019}]%
        {Microsoft}
 \bibinfo{year}{2019}\natexlab{}.
\newblock \bibinfo{title}{Microsoft InterpretML}.
\newblock
\newblock
\newblock
\shownote{hhttps://github.com/interpretml/interpret.}


\bibitem[\protect\citeauthoryear{Abdul, von~der Weth, Kankanhalli, and
  Lim}{Abdul et~al\mbox{.}}{2020}]%
        {abdul2020cogam}
\bibfield{author}{\bibinfo{person}{Ashraf Abdul}, \bibinfo{person}{Christian
  von~der Weth}, \bibinfo{person}{Mohan Kankanhalli}, {and}
  \bibinfo{person}{Brian~Y Lim}.} \bibinfo{year}{2020}\natexlab{}.
\newblock \showarticletitle{COGAM: Measuring and Moderating Cognitive Load in
  Machine Learning Model Explanations}. In
  \bibinfo{booktitle}{\emph{Proceedings of the 2020 CHI Conference on Human
  Factors in Computing Systems}}. \bibinfo{pages}{1--14}.
\newblock


\bibitem[\protect\citeauthoryear{Adadi and Berrada}{Adadi and Berrada}{2018}]%
        {adadi2018peeking}
\bibfield{author}{\bibinfo{person}{Amina Adadi} {and} \bibinfo{person}{Mohammed
  Berrada}.} \bibinfo{year}{2018}\natexlab{}.
\newblock \showarticletitle{Peeking inside the black-box: A survey on
  Explainable Artificial Intelligence (XAI)}.
\newblock \bibinfo{journal}{\emph{IEEE Access}}  \bibinfo{volume}{6}
  (\bibinfo{year}{2018}), \bibinfo{pages}{52138--52160}.
\newblock


\bibitem[\protect\citeauthoryear{Alvarez-Melis and Jaakkola}{Alvarez-Melis and
  Jaakkola}{2018}]%
        {alvarez2018towards}
\bibfield{author}{\bibinfo{person}{David Alvarez-Melis} {and}
  \bibinfo{person}{Tommi~S Jaakkola}.} \bibinfo{year}{2018}\natexlab{}.
\newblock \showarticletitle{Towards robust interpretability with
  self-explaining neural networks}.
\newblock \bibinfo{journal}{\emph{arXiv preprint arXiv:1806.07538}}
  (\bibinfo{year}{2018}).
\newblock


\bibitem[\protect\citeauthoryear{Alvarez-Melis, Kaur, Daum{\'e}~III, Wallach,
  and Vaughan}{Alvarez-Melis et~al\mbox{.}}{2021}]%
        {alvarez2021human}
\bibfield{author}{\bibinfo{person}{David Alvarez-Melis},
  \bibinfo{person}{Harmanpreet Kaur}, \bibinfo{person}{Hal Daum{\'e}~III},
  \bibinfo{person}{Hanna Wallach}, {and} \bibinfo{person}{Jennifer~Wortman
  Vaughan}.} \bibinfo{year}{2021}\natexlab{}.
\newblock \showarticletitle{From human explanation to model interpretability: A
  framework based on weight of evidence}. In \bibinfo{booktitle}{\emph{AAAI
  Conference on Human Computation and Crowdsourcing (HCOMP)}}.
\newblock


\bibitem[\protect\citeauthoryear{Antaki and Leudar}{Antaki and Leudar}{1992}]%
        {antaki1992explaining}
\bibfield{author}{\bibinfo{person}{Charles Antaki} {and} \bibinfo{person}{Ivan
  Leudar}.} \bibinfo{year}{1992}\natexlab{}.
\newblock \showarticletitle{Explaining in conversation: Towards an argument
  model}.
\newblock \bibinfo{journal}{\emph{European Journal of Social Psychology}}
  \bibinfo{volume}{22}, \bibinfo{number}{2} (\bibinfo{year}{1992}),
  \bibinfo{pages}{181--194}.
\newblock


\bibitem[\protect\citeauthoryear{Apley and Zhu}{Apley and Zhu}{2020}]%
        {apley2020visualizing}
\bibfield{author}{\bibinfo{person}{Daniel~W Apley} {and}
  \bibinfo{person}{Jingyu Zhu}.} \bibinfo{year}{2020}\natexlab{}.
\newblock \showarticletitle{Visualizing the effects of predictor variables in
  black box supervised learning models}.
\newblock \bibinfo{journal}{\emph{Journal of the Royal Statistical Society:
  Series B (Statistical Methodology)}} \bibinfo{volume}{82},
  \bibinfo{number}{4} (\bibinfo{year}{2020}), \bibinfo{pages}{1059--1086}.
\newblock


\bibitem[\protect\citeauthoryear{Arnold, Bellamy, Hind, Houde, Mehta,
  Mojsilovi{\'c}, Nair, Ramamurthy, Olteanu, Piorkowski, et~al\mbox{.}}{Arnold
  et~al\mbox{.}}{2019}]%
        {arnold2019factsheets}
\bibfield{author}{\bibinfo{person}{Matthew Arnold}, \bibinfo{person}{Rachel~KE
  Bellamy}, \bibinfo{person}{Michael Hind}, \bibinfo{person}{Stephanie Houde},
  \bibinfo{person}{Sameep Mehta}, \bibinfo{person}{Aleksandra Mojsilovi{\'c}},
  \bibinfo{person}{Ravi Nair}, \bibinfo{person}{K~Natesan Ramamurthy},
  \bibinfo{person}{Alexandra Olteanu}, \bibinfo{person}{David Piorkowski},
  {et~al\mbox{.}}} \bibinfo{year}{2019}\natexlab{}.
\newblock \showarticletitle{FactSheets: Increasing trust in AI services through
  supplier's declarations of conformity}.
\newblock \bibinfo{journal}{\emph{IBM Journal of Research and Development}}
  \bibinfo{volume}{63}, \bibinfo{number}{4/5} (\bibinfo{year}{2019}),
  \bibinfo{pages}{6--1}.
\newblock


\bibitem[\protect\citeauthoryear{Arrieta, D{\'\i}az-Rodr{\'\i}guez, Del~Ser,
  Bennetot, Tabik, Barbado, Garc{\'\i}a, Gil-L{\'o}pez, Molina, Benjamins,
  et~al\mbox{.}}{Arrieta et~al\mbox{.}}{2020}]%
        {arrieta2020explainable}
\bibfield{author}{\bibinfo{person}{Alejandro~Barredo Arrieta},
  \bibinfo{person}{Natalia D{\'\i}az-Rodr{\'\i}guez}, \bibinfo{person}{Javier
  Del~Ser}, \bibinfo{person}{Adrien Bennetot}, \bibinfo{person}{Siham Tabik},
  \bibinfo{person}{Alberto Barbado}, \bibinfo{person}{Salvador Garc{\'\i}a},
  \bibinfo{person}{Sergio Gil-L{\'o}pez}, \bibinfo{person}{Daniel Molina},
  \bibinfo{person}{Richard Benjamins}, {et~al\mbox{.}}}
  \bibinfo{year}{2020}\natexlab{}.
\newblock \showarticletitle{Explainable Artificial Intelligence (XAI):
  Concepts, taxonomies, opportunities and challenges toward responsible AI}.
\newblock \bibinfo{journal}{\emph{Information Fusion}}  \bibinfo{volume}{58}
  (\bibinfo{year}{2020}), \bibinfo{pages}{82--115}.
\newblock


\bibitem[\protect\citeauthoryear{Arya, Bellamy, Chen, Dhurandhar, Hind,
  Hoffman, Houde, Liao, Luss, Mojsilovic, Mourad, Pedemonte, Raghavendra,
  Richards, Sattigeri, Shanmugam, Singh, Varshney, Wei, and Zhang}{Arya
  et~al\mbox{.}}{2020}]%
        {arya2020ai}
\bibfield{author}{\bibinfo{person}{Vijay Arya}, \bibinfo{person}{Rachel K.~E.
  Bellamy}, \bibinfo{person}{Pin-Yu Chen}, \bibinfo{person}{Amit Dhurandhar},
  \bibinfo{person}{Michael Hind}, \bibinfo{person}{Samuel~C. Hoffman},
  \bibinfo{person}{Stephanie Houde}, \bibinfo{person}{Q.~Vera Liao},
  \bibinfo{person}{Ronny Luss}, \bibinfo{person}{Aleksandra Mojsilovic},
  \bibinfo{person}{Sami Mourad}, \bibinfo{person}{Pablo Pedemonte},
  \bibinfo{person}{Ramya Raghavendra}, \bibinfo{person}{John Richards},
  \bibinfo{person}{Prasanna Sattigeri}, \bibinfo{person}{Karthikeyan
  Shanmugam}, \bibinfo{person}{Moninder Singh}, \bibinfo{person}{Kush~R.
  Varshney}, \bibinfo{person}{Dennis Wei}, {and} \bibinfo{person}{Yunfeng
  Zhang}.} \bibinfo{year}{2020}\natexlab{}.
\newblock \showarticletitle{{AI} {E}xplainability 360: An Extensible Toolkit
  for Understanding Data and Machine Learning Models}.
\newblock \bibinfo{journal}{\emph{J. Mach. Learn. Res.}} \bibinfo{volume}{21},
  \bibinfo{number}{130} (\bibinfo{year}{2020}), \bibinfo{pages}{1--6}.
\newblock


\bibitem[\protect\citeauthoryear{Bach, Binder, Montavon, Klauschen, M{\"u}ller,
  and Samek}{Bach et~al\mbox{.}}{2015}]%
        {bach2015pixel}
\bibfield{author}{\bibinfo{person}{Sebastian Bach}, \bibinfo{person}{Alexander
  Binder}, \bibinfo{person}{Gr{\'e}goire Montavon}, \bibinfo{person}{Frederick
  Klauschen}, \bibinfo{person}{Klaus-Robert M{\"u}ller}, {and}
  \bibinfo{person}{Wojciech Samek}.} \bibinfo{year}{2015}\natexlab{}.
\newblock \showarticletitle{On pixel-wise explanations for non-linear
  classifier decisions by layer-wise relevance propagation}.
\newblock \bibinfo{journal}{\emph{PloS one}} \bibinfo{volume}{10},
  \bibinfo{number}{7} (\bibinfo{year}{2015}), \bibinfo{pages}{e0130140}.
\newblock


\bibitem[\protect\citeauthoryear{Bansal, Wu, Zhou, Fok, Nushi, Kamar, Ribeiro,
  and Weld}{Bansal et~al\mbox{.}}{2021}]%
        {bansal2021does}
\bibfield{author}{\bibinfo{person}{Gagan Bansal}, \bibinfo{person}{Tongshuang
  Wu}, \bibinfo{person}{Joyce Zhou}, \bibinfo{person}{Raymond Fok},
  \bibinfo{person}{Besmira Nushi}, \bibinfo{person}{Ece Kamar},
  \bibinfo{person}{Marco~Tulio Ribeiro}, {and} \bibinfo{person}{Daniel Weld}.}
  \bibinfo{year}{2021}\natexlab{}.
\newblock \showarticletitle{Does the whole exceed its parts? the effect of ai
  explanations on complementary team performance}. In
  \bibinfo{booktitle}{\emph{Proceedings of the 2021 CHI Conference on Human
  Factors in Computing Systems}}. \bibinfo{pages}{1--16}.
\newblock


\bibitem[\protect\citeauthoryear{Bickmore and Cassell}{Bickmore and
  Cassell}{2001}]%
        {bickmore2001relational}
\bibfield{author}{\bibinfo{person}{Timothy Bickmore} {and}
  \bibinfo{person}{Justine Cassell}.} \bibinfo{year}{2001}\natexlab{}.
\newblock \showarticletitle{Relational agents: a model and implementation of
  building user trust}. In \bibinfo{booktitle}{\emph{Proceedings of the SIGCHI
  conference on Human factors in computing systems}}.
  \bibinfo{pages}{396--403}.
\newblock


\bibitem[\protect\citeauthoryear{Bu{\c{c}}inca, Lin, Gajos, and
  Glassman}{Bu{\c{c}}inca et~al\mbox{.}}{2020}]%
        {buccinca2020proxy}
\bibfield{author}{\bibinfo{person}{Zana Bu{\c{c}}inca}, \bibinfo{person}{Phoebe
  Lin}, \bibinfo{person}{Krzysztof~Z Gajos}, {and} \bibinfo{person}{Elena~L
  Glassman}.} \bibinfo{year}{2020}\natexlab{}.
\newblock \showarticletitle{Proxy tasks and subjective measures can be
  misleading in evaluating explainable AI systems}. In
  \bibinfo{booktitle}{\emph{Proceedings of the 25th International Conference on
  Intelligent User Interfaces}}. \bibinfo{pages}{454--464}.
\newblock


\bibitem[\protect\citeauthoryear{Bu{\c{c}}inca, Malaya, and
  Gajos}{Bu{\c{c}}inca et~al\mbox{.}}{2021}]%
        {buccinca2021trust}
\bibfield{author}{\bibinfo{person}{Zana Bu{\c{c}}inca},
  \bibinfo{person}{Maja~Barbara Malaya}, {and} \bibinfo{person}{Krzysztof~Z
  Gajos}.} \bibinfo{year}{2021}\natexlab{}.
\newblock \showarticletitle{To trust or to think: cognitive forcing functions
  can reduce overreliance on AI in AI-assisted decision-making}.
\newblock \bibinfo{journal}{\emph{Proceedings of the ACM on Human-Computer
  Interaction}} \bibinfo{volume}{5}, \bibinfo{number}{CSCW1}
  (\bibinfo{year}{2021}), \bibinfo{pages}{1--21}.
\newblock


\bibitem[\protect\citeauthoryear{Cai, Winter, Steiner, Wilcox, and Terry}{Cai
  et~al\mbox{.}}{2019}]%
        {cai2019hello}
\bibfield{author}{\bibinfo{person}{Carrie~J Cai}, \bibinfo{person}{Samantha
  Winter}, \bibinfo{person}{David Steiner}, \bibinfo{person}{Lauren Wilcox},
  {and} \bibinfo{person}{Michael Terry}.} \bibinfo{year}{2019}\natexlab{}.
\newblock \showarticletitle{Hello AI: Uncovering the Onboarding Needs of
  Medical Practitioners for Human-AI Collaborative Decision-Making}.
\newblock \bibinfo{journal}{\emph{Proceedings of the ACM on Human-Computer
  Interaction}} \bibinfo{volume}{3}, \bibinfo{number}{CSCW}
  (\bibinfo{year}{2019}), \bibinfo{pages}{104}.
\newblock


\bibitem[\protect\citeauthoryear{Caruana, Lou, Gehrke, Koch, Sturm, and
  Elhadad}{Caruana et~al\mbox{.}}{2015}]%
        {caruana2015intelligible}
\bibfield{author}{\bibinfo{person}{Rich Caruana}, \bibinfo{person}{Yin Lou},
  \bibinfo{person}{Johannes Gehrke}, \bibinfo{person}{Paul Koch},
  \bibinfo{person}{Marc Sturm}, {and} \bibinfo{person}{Noemie Elhadad}.}
  \bibinfo{year}{2015}\natexlab{}.
\newblock \showarticletitle{Intelligible models for healthcare: Predicting
  pneumonia risk and hospital 30-day readmission}. In
  \bibinfo{booktitle}{\emph{Proceedings of KDD}}.
\newblock


\bibitem[\protect\citeauthoryear{Carvalho, Pereira, and Cardoso}{Carvalho
  et~al\mbox{.}}{2019}]%
        {carvalho2019machine}
\bibfield{author}{\bibinfo{person}{Diogo~V Carvalho},
  \bibinfo{person}{Eduardo~M Pereira}, {and} \bibinfo{person}{Jaime~S
  Cardoso}.} \bibinfo{year}{2019}\natexlab{}.
\newblock \showarticletitle{Machine learning interpretability: A survey on
  methods and metrics}.
\newblock \bibinfo{journal}{\emph{Electronics}} \bibinfo{volume}{8},
  \bibinfo{number}{8} (\bibinfo{year}{2019}), \bibinfo{pages}{832}.
\newblock


\bibitem[\protect\citeauthoryear{Cawsey}{Cawsey}{1992}]%
        {cawsey1992explanation}
\bibfield{author}{\bibinfo{person}{Alison Cawsey}.}
  \bibinfo{year}{1992}\natexlab{}.
\newblock \bibinfo{booktitle}{\emph{Explanation and interaction: the computer
  generation of explanatory dialogues}}.
\newblock \bibinfo{publisher}{MIT press}.
\newblock


\bibitem[\protect\citeauthoryear{Chromik, Eiband, Buchner, Kr{\"u}ger, and
  Butz}{Chromik et~al\mbox{.}}{2021}]%
        {chromik2021think}
\bibfield{author}{\bibinfo{person}{Michael Chromik}, \bibinfo{person}{Malin
  Eiband}, \bibinfo{person}{Felicitas Buchner}, \bibinfo{person}{Adrian
  Kr{\"u}ger}, {and} \bibinfo{person}{Andreas Butz}.}
  \bibinfo{year}{2021}\natexlab{}.
\newblock \showarticletitle{I Think I Get Your Point, AI! The Illusion of
  Explanatory Depth in Explainable AI}. In \bibinfo{booktitle}{\emph{26th
  International Conference on Intelligent User Interfaces}}.
  \bibinfo{pages}{307--317}.
\newblock


\bibitem[\protect\citeauthoryear{Clark and Brennan}{Clark and Brennan}{1991}]%
        {clark1991grounding}
\bibfield{author}{\bibinfo{person}{Herbert~H Clark} {and}
  \bibinfo{person}{Susan~E Brennan}.} \bibinfo{year}{1991}\natexlab{}.
\newblock \showarticletitle{Grounding in communication.}
\newblock  (\bibinfo{year}{1991}).
\newblock


\bibitem[\protect\citeauthoryear{Craven and Shavlik}{Craven and
  Shavlik}{1995}]%
        {craven1995extracting}
\bibfield{author}{\bibinfo{person}{Mark Craven} {and} \bibinfo{person}{Jude
  Shavlik}.} \bibinfo{year}{1995}\natexlab{}.
\newblock \showarticletitle{Extracting tree-structured representations of
  trained networks}.
\newblock \bibinfo{journal}{\emph{Advances in neural information processing
  systems}}  \bibinfo{volume}{8} (\bibinfo{year}{1995}),
  \bibinfo{pages}{24--30}.
\newblock


\bibitem[\protect\citeauthoryear{Dervin}{Dervin}{1998}]%
        {dervin1998sense}
\bibfield{author}{\bibinfo{person}{Brenda Dervin}.}
  \bibinfo{year}{1998}\natexlab{}.
\newblock \showarticletitle{Sense-making theory and practice: An overview of
  user interests in knowledge seeking and use}.
\newblock \bibinfo{journal}{\emph{Journal of knowledge management}}
  (\bibinfo{year}{1998}).
\newblock


\bibitem[\protect\citeauthoryear{Dhurandhar, Chen, Luss, Tu, Ting, Shanmugam,
  and Das}{Dhurandhar et~al\mbox{.}}{2018a}]%
        {dhurandhar2018explanations}
\bibfield{author}{\bibinfo{person}{Amit Dhurandhar}, \bibinfo{person}{Pin-Yu
  Chen}, \bibinfo{person}{Ronny Luss}, \bibinfo{person}{Chun-Chen Tu},
  \bibinfo{person}{Paishun Ting}, \bibinfo{person}{Karthikeyan Shanmugam},
  {and} \bibinfo{person}{Payel Das}.} \bibinfo{year}{2018}\natexlab{a}.
\newblock \showarticletitle{Explanations based on the missing: Towards
  contrastive explanations with pertinent negatives}.
\newblock \bibinfo{journal}{\emph{arXiv preprint arXiv:1802.07623}}
  (\bibinfo{year}{2018}).
\newblock


\bibitem[\protect\citeauthoryear{Dhurandhar, Shanmugam, Luss, and
  Olsen}{Dhurandhar et~al\mbox{.}}{2018b}]%
        {dhurandhar2018improving}
\bibfield{author}{\bibinfo{person}{Amit Dhurandhar},
  \bibinfo{person}{Karthikeyan Shanmugam}, \bibinfo{person}{Ronny Luss}, {and}
  \bibinfo{person}{Peder~A Olsen}.} \bibinfo{year}{2018}\natexlab{b}.
\newblock \showarticletitle{Improving Simple Models with Confidence Profiles}.
\newblock \bibinfo{journal}{\emph{Advances in Neural Information Processing
  Systems}}  \bibinfo{volume}{31} (\bibinfo{year}{2018}).
\newblock


\bibitem[\protect\citeauthoryear{Dodge, Liao, Zhang, Bellamy, and Dugan}{Dodge
  et~al\mbox{.}}{2019}]%
        {dodge2019explaining}
\bibfield{author}{\bibinfo{person}{Jonathan Dodge}, \bibinfo{person}{Q~Vera
  Liao}, \bibinfo{person}{Yunfeng Zhang}, \bibinfo{person}{Rachel~KE Bellamy},
  {and} \bibinfo{person}{Casey Dugan}.} \bibinfo{year}{2019}\natexlab{}.
\newblock \showarticletitle{Explaining models: an empirical study of how
  explanations impact fairness judgment}. In
  \bibinfo{booktitle}{\emph{Proceedings of the 24th International Conference on
  Intelligent User Interfaces}}. \bibinfo{pages}{275--285}.
\newblock


\bibitem[\protect\citeauthoryear{Doshi-Velez and Kim}{Doshi-Velez and
  Kim}{2017}]%
        {doshi2017towards}
\bibfield{author}{\bibinfo{person}{Finale Doshi-Velez} {and}
  \bibinfo{person}{Been Kim}.} \bibinfo{year}{2017}\natexlab{}.
\newblock \showarticletitle{Towards a rigorous science of interpretable machine
  learning}.
\newblock \bibinfo{journal}{\emph{arXiv preprint arXiv:1702.08608}}
  (\bibinfo{year}{2017}).
\newblock


\bibitem[\protect\citeauthoryear{Ehsan, Liao, Muller, Riedl, and Weisz}{Ehsan
  et~al\mbox{.}}{2021a}]%
        {ehsan2021expanding}
\bibfield{author}{\bibinfo{person}{Upol Ehsan}, \bibinfo{person}{Q~Vera Liao},
  \bibinfo{person}{Michael Muller}, \bibinfo{person}{Mark~O Riedl}, {and}
  \bibinfo{person}{Justin~D Weisz}.} \bibinfo{year}{2021}\natexlab{a}.
\newblock \showarticletitle{Expanding explainability: Towards social
  transparency in ai systems}. In \bibinfo{booktitle}{\emph{Proceedings of the
  2021 CHI Conference on Human Factors in Computing Systems}}.
  \bibinfo{pages}{1--19}.
\newblock


\bibitem[\protect\citeauthoryear{Ehsan, Passi, Liao, Chan, Lee, Muller, Riedl,
  et~al\mbox{.}}{Ehsan et~al\mbox{.}}{2021b}]%
        {ehsan2021explainable}
\bibfield{author}{\bibinfo{person}{Upol Ehsan}, \bibinfo{person}{Samir Passi},
  \bibinfo{person}{Q~Vera Liao}, \bibinfo{person}{Larry Chan},
  \bibinfo{person}{I Lee}, \bibinfo{person}{Michael Muller},
  \bibinfo{person}{Mark~O Riedl}, {et~al\mbox{.}}}
  \bibinfo{year}{2021}\natexlab{b}.
\newblock \showarticletitle{The Who in Explainable AI: How AI Background Shapes
  Perceptions of AI Explanations}.
\newblock \bibinfo{journal}{\emph{arXiv preprint arXiv:2107.13509}}
  (\bibinfo{year}{2021}).
\newblock


\bibitem[\protect\citeauthoryear{Ehsan and Riedl}{Ehsan and Riedl}{2020}]%
        {ehsan2020human}
\bibfield{author}{\bibinfo{person}{Upol Ehsan} {and} \bibinfo{person}{Mark~O
  Riedl}.} \bibinfo{year}{2020}\natexlab{}.
\newblock \showarticletitle{Human-centered explainable ai: Towards a reflective
  sociotechnical approach}. In \bibinfo{booktitle}{\emph{International
  Conference on Human-Computer Interaction}}. Springer,
  \bibinfo{pages}{449--466}.
\newblock


\bibitem[\protect\citeauthoryear{Ehsan, Tambwekar, Chan, Harrison, and
  Riedl}{Ehsan et~al\mbox{.}}{2019}]%
        {ehsan2019automated}
\bibfield{author}{\bibinfo{person}{Upol Ehsan}, \bibinfo{person}{Pradyumna
  Tambwekar}, \bibinfo{person}{Larry Chan}, \bibinfo{person}{Brent Harrison},
  {and} \bibinfo{person}{Mark~O Riedl}.} \bibinfo{year}{2019}\natexlab{}.
\newblock \showarticletitle{Automated rationale generation: a technique for
  explainable AI and its effects on human perceptions}. In
  \bibinfo{booktitle}{\emph{Proceedings of the 24th International Conference on
  Intelligent User Interfaces}}. \bibinfo{pages}{263--274}.
\newblock


\bibitem[\protect\citeauthoryear{Ehsan, Wintersberger, Liao, Mara, Streit,
  Wachter, Riener, and Riedl}{Ehsan et~al\mbox{.}}{2021c}]%
        {ehsan2021operationalizing}
\bibfield{author}{\bibinfo{person}{Upol Ehsan}, \bibinfo{person}{Philipp
  Wintersberger}, \bibinfo{person}{Q~Vera Liao}, \bibinfo{person}{Martina
  Mara}, \bibinfo{person}{Marc Streit}, \bibinfo{person}{Sandra Wachter},
  \bibinfo{person}{Andreas Riener}, {and} \bibinfo{person}{Mark~O Riedl}.}
  \bibinfo{year}{2021}\natexlab{c}.
\newblock \showarticletitle{Operationalizing Human-Centered Perspectives in
  Explainable AI}. In \bibinfo{booktitle}{\emph{Extended Abstracts of the 2021
  CHI Conference on Human Factors in Computing Systems}}.
  \bibinfo{pages}{1--6}.
\newblock


\bibitem[\protect\citeauthoryear{Eiband, Buschek, Kremer, and Hussmann}{Eiband
  et~al\mbox{.}}{2019}]%
        {eiband2019impact}
\bibfield{author}{\bibinfo{person}{Malin Eiband}, \bibinfo{person}{Daniel
  Buschek}, \bibinfo{person}{Alexander Kremer}, {and} \bibinfo{person}{Heinrich
  Hussmann}.} \bibinfo{year}{2019}\natexlab{}.
\newblock \showarticletitle{The impact of placebic explanations on trust in
  intelligent systems}. In \bibinfo{booktitle}{\emph{Extended Abstracts of the
  2019 CHI Conference on Human Factors in Computing Systems}}.
  \bibinfo{pages}{1--6}.
\newblock


\bibitem[\protect\citeauthoryear{Eiband, Schneider, Bilandzic, Fazekas-Con,
  Haug, and Hussmann}{Eiband et~al\mbox{.}}{2018}]%
        {eiband2018bringing}
\bibfield{author}{\bibinfo{person}{Malin Eiband}, \bibinfo{person}{Hanna
  Schneider}, \bibinfo{person}{Mark Bilandzic}, \bibinfo{person}{Julian
  Fazekas-Con}, \bibinfo{person}{Mareike Haug}, {and} \bibinfo{person}{Heinrich
  Hussmann}.} \bibinfo{year}{2018}\natexlab{}.
\newblock \showarticletitle{Bringing transparency design into practice}. In
  \bibinfo{booktitle}{\emph{23rd international conference on intelligent user
  interfaces}}. \bibinfo{pages}{211--223}.
\newblock


\bibitem[\protect\citeauthoryear{Fan, Yang, Yu, Liao, and Zhao}{Fan
  et~al\mbox{.}}{2022}]%
        {fan2022human}
\bibfield{author}{\bibinfo{person}{Mingming Fan}, \bibinfo{person}{Xianyou
  Yang}, \bibinfo{person}{TszTung Yu}, \bibinfo{person}{Q~Vera Liao}, {and}
  \bibinfo{person}{Jian Zhao}.} \bibinfo{year}{2022}\natexlab{}.
\newblock \showarticletitle{Human-AI Collaboration for UX Evaluation: Effects
  of Explanation and Synchronization}.
\newblock \bibinfo{journal}{\emph{Proceedings of the ACM on Human-Computer
  Interaction}} \bibinfo{volume}{6}, \bibinfo{number}{CSCW1}
  (\bibinfo{year}{2022}), \bibinfo{pages}{1--32}.
\newblock


\bibitem[\protect\citeauthoryear{Gebru, Morgenstern, Vecchione,
  Wortman~Vaughan, Wallach, Daum{\'e}, and Crawford}{Gebru
  et~al\mbox{.}}{2021}]%
        {gebru2021datasheets}
\bibfield{author}{\bibinfo{person}{Timnit Gebru}, \bibinfo{person}{Jamie
  Morgenstern}, \bibinfo{person}{Briana Vecchione}, \bibinfo{person}{Jennifer
  Wortman~Vaughan}, \bibinfo{person}{Hanna Wallach}, \bibinfo{person}{Hal
  Daum{\'e}, {III}}, {and} \bibinfo{person}{Kate Crawford}.}
  \bibinfo{year}{2021}\natexlab{}.
\newblock \showarticletitle{Datasheets for Datasets}.
\newblock \bibinfo{journal}{\emph{Commun. ACM}} \bibinfo{volume}{64},
  \bibinfo{number}{12} (\bibinfo{year}{2021}), \bibinfo{pages}{86--92}.
\newblock


\bibitem[\protect\citeauthoryear{Ghai, Liao, Zhang, Bellamy, and Mueller}{Ghai
  et~al\mbox{.}}{2021}]%
        {ghai2021explainable}
\bibfield{author}{\bibinfo{person}{Bhavya Ghai}, \bibinfo{person}{Q~Vera Liao},
  \bibinfo{person}{Yunfeng Zhang}, \bibinfo{person}{Rachel Bellamy}, {and}
  \bibinfo{person}{Klaus Mueller}.} \bibinfo{year}{2021}\natexlab{}.
\newblock \showarticletitle{Explainable active learning (xal) toward ai
  explanations as interfaces for machine teachers}.
\newblock \bibinfo{journal}{\emph{Proceedings of the ACM on Human-Computer
  Interaction}} \bibinfo{volume}{4}, \bibinfo{number}{CSCW3}
  (\bibinfo{year}{2021}), \bibinfo{pages}{1--28}.
\newblock


\bibitem[\protect\citeauthoryear{Ghassemi, Oakden-Rayner, and Beam}{Ghassemi
  et~al\mbox{.}}{2021}]%
        {ghassemi2021false}
\bibfield{author}{\bibinfo{person}{Marzyeh Ghassemi}, \bibinfo{person}{Luke
  Oakden-Rayner}, {and} \bibinfo{person}{Andrew~L Beam}.}
  \bibinfo{year}{2021}\natexlab{}.
\newblock \showarticletitle{The false hope of current approaches to explainable
  artificial intelligence in health care}.
\newblock \bibinfo{journal}{\emph{The Lancet Digital Health}}
  \bibinfo{volume}{3}, \bibinfo{number}{11} (\bibinfo{year}{2021}),
  \bibinfo{pages}{e745--e750}.
\newblock


\bibitem[\protect\citeauthoryear{Ghosh, Liao, Ramamurthy, Navratil, Sattigeri,
  Varshney, and Zhang}{Ghosh et~al\mbox{.}}{2021}]%
        {ghosh2021uncertainty}
\bibfield{author}{\bibinfo{person}{Soumya Ghosh}, \bibinfo{person}{Q~Vera
  Liao}, \bibinfo{person}{Karthikeyan~Natesan Ramamurthy},
  \bibinfo{person}{Jiri Navratil}, \bibinfo{person}{Prasanna Sattigeri},
  \bibinfo{person}{Kush~R Varshney}, {and} \bibinfo{person}{Yunfeng Zhang}.}
  \bibinfo{year}{2021}\natexlab{}.
\newblock \showarticletitle{Uncertainty Quantification 360: A Holistic Toolkit
  for Quantifying and Communicating the Uncertainty of AI}.
\newblock \bibinfo{journal}{\emph{arXiv preprint arXiv:2106.01410}}
  (\bibinfo{year}{2021}).
\newblock


\bibitem[\protect\citeauthoryear{Goffman et~al\mbox{.}}{Goffman
  et~al\mbox{.}}{1978}]%
        {goffman1978presentation}
\bibfield{author}{\bibinfo{person}{Erving Goffman} {et~al\mbox{.}}}
  \bibinfo{year}{1978}\natexlab{}.
\newblock \bibinfo{booktitle}{\emph{The presentation of self in everyday
  life}}. Vol.~\bibinfo{volume}{21}.
\newblock \bibinfo{publisher}{Harmondsworth London}.
\newblock


\bibitem[\protect\citeauthoryear{Goldstein, Kapelner, Bleich, and
  Pitkin}{Goldstein et~al\mbox{.}}{2015}]%
        {goldstein2015peeking}
\bibfield{author}{\bibinfo{person}{Alex Goldstein}, \bibinfo{person}{Adam
  Kapelner}, \bibinfo{person}{Justin Bleich}, {and} \bibinfo{person}{Emil
  Pitkin}.} \bibinfo{year}{2015}\natexlab{}.
\newblock \showarticletitle{Peeking inside the black box: Visualizing
  statistical learning with plots of individual conditional expectation}.
\newblock \bibinfo{journal}{\emph{journal of Computational and Graphical
  Statistics}} \bibinfo{volume}{24}, \bibinfo{number}{1}
  (\bibinfo{year}{2015}), \bibinfo{pages}{44--65}.
\newblock


\bibitem[\protect\citeauthoryear{Grice}{Grice}{1975}]%
        {grice1975logic}
\bibfield{author}{\bibinfo{person}{Herbert~P Grice}.}
  \bibinfo{year}{1975}\natexlab{}.
\newblock \showarticletitle{Logic and conversation}.
\newblock In \bibinfo{booktitle}{\emph{Speech acts}}.
  \bibinfo{publisher}{Brill}, \bibinfo{pages}{41--58}.
\newblock


\bibitem[\protect\citeauthoryear{Guidotti, Monreale, Ruggieri, Turini,
  Giannotti, and Pedreschi}{Guidotti et~al\mbox{.}}{2019}]%
        {guidotti2019survey}
\bibfield{author}{\bibinfo{person}{Riccardo Guidotti}, \bibinfo{person}{Anna
  Monreale}, \bibinfo{person}{Salvatore Ruggieri}, \bibinfo{person}{Franco
  Turini}, \bibinfo{person}{Fosca Giannotti}, {and} \bibinfo{person}{Dino
  Pedreschi}.} \bibinfo{year}{2019}\natexlab{}.
\newblock \showarticletitle{A survey of methods for explaining black box
  models}.
\newblock \bibinfo{journal}{\emph{ACM computing surveys (CSUR)}}
  \bibinfo{volume}{51}, \bibinfo{number}{5} (\bibinfo{year}{2019}),
  \bibinfo{pages}{93}.
\newblock


\bibitem[\protect\citeauthoryear{Gurumoorthy, Dhurandhar, Cecchi, and
  Aggarwal}{Gurumoorthy et~al\mbox{.}}{2019}]%
        {gurumoorthy2019efficient}
\bibfield{author}{\bibinfo{person}{Karthik~S Gurumoorthy},
  \bibinfo{person}{Amit Dhurandhar}, \bibinfo{person}{Guillermo Cecchi}, {and}
  \bibinfo{person}{Charu Aggarwal}.} \bibinfo{year}{2019}\natexlab{}.
\newblock \showarticletitle{Efficient data representation by selecting
  prototypes with importance weights}. In \bibinfo{booktitle}{\emph{2019 IEEE
  International Conference on Data Mining (ICDM)}}. IEEE,
  \bibinfo{pages}{260--269}.
\newblock


\bibitem[\protect\citeauthoryear{Hase and Bansal}{Hase and Bansal}{2020}]%
        {hase2020evaluating}
\bibfield{author}{\bibinfo{person}{Peter Hase} {and} \bibinfo{person}{Mohit
  Bansal}.} \bibinfo{year}{2020}\natexlab{}.
\newblock \showarticletitle{Evaluating Explainable AI: Which Algorithmic
  Explanations Help Users Predict Model Behavior?}. In
  \bibinfo{booktitle}{\emph{Proceedings of the 58th Annual Meeting of the
  Association for Computational Linguistics}}. \bibinfo{pages}{5540--5552}.
\newblock


\bibitem[\protect\citeauthoryear{Hastie, Tibshirani, and Friedman}{Hastie
  et~al\mbox{.}}{2009}]%
        {hastie2009elements}
\bibfield{author}{\bibinfo{person}{Trevor Hastie}, \bibinfo{person}{Robert
  Tibshirani}, {and} \bibinfo{person}{Jerome Friedman}.}
  \bibinfo{year}{2009}\natexlab{}.
\newblock \showarticletitle{The elements of statistical learnin}.
\newblock \bibinfo{journal}{\emph{Cited on}} (\bibinfo{year}{2009}),
  \bibinfo{pages}{33}.
\newblock


\bibitem[\protect\citeauthoryear{Hilton}{Hilton}{1990}]%
        {hilton1990conversational}
\bibfield{author}{\bibinfo{person}{Denis~J Hilton}.}
  \bibinfo{year}{1990}\natexlab{}.
\newblock \showarticletitle{Conversational processes and causal explanation.}
\newblock \bibinfo{journal}{\emph{Psychological Bulletin}}
  \bibinfo{volume}{107}, \bibinfo{number}{1} (\bibinfo{year}{1990}),
  \bibinfo{pages}{65}.
\newblock


\bibitem[\protect\citeauthoryear{Hind}{Hind}{2019}]%
        {hind2019explaining}
\bibfield{author}{\bibinfo{person}{Michael Hind}.}
  \bibinfo{year}{2019}\natexlab{}.
\newblock \showarticletitle{Explaining explainable AI}.
\newblock \bibinfo{journal}{\emph{XRDS: Crossroads, The ACM Magazine for
  Students}} \bibinfo{volume}{25}, \bibinfo{number}{3} (\bibinfo{year}{2019}),
  \bibinfo{pages}{16--19}.
\newblock


\bibitem[\protect\citeauthoryear{Hind, Wei, Campbell, Codella, Dhurandhar,
  Mojsilovi{\'c}, Natesan~Ramamurthy, and Varshney}{Hind et~al\mbox{.}}{2019}]%
        {hind2019ted}
\bibfield{author}{\bibinfo{person}{Michael Hind}, \bibinfo{person}{Dennis Wei},
  \bibinfo{person}{Murray Campbell}, \bibinfo{person}{Noel~CF Codella},
  \bibinfo{person}{Amit Dhurandhar}, \bibinfo{person}{Aleksandra
  Mojsilovi{\'c}}, \bibinfo{person}{Karthikeyan Natesan~Ramamurthy}, {and}
  \bibinfo{person}{Kush~R Varshney}.} \bibinfo{year}{2019}\natexlab{}.
\newblock \showarticletitle{TED: Teaching AI to explain its decisions}. In
  \bibinfo{booktitle}{\emph{Proceedings of the 2019 AAAI/ACM Conference on AI,
  Ethics, and Society}}. \bibinfo{pages}{123--129}.
\newblock


\bibitem[\protect\citeauthoryear{Hohman, Head, Caruana, DeLine, and
  Drucker}{Hohman et~al\mbox{.}}{2019}]%
        {hohman2019gamut}
\bibfield{author}{\bibinfo{person}{Fred Hohman}, \bibinfo{person}{Andrew Head},
  \bibinfo{person}{Rich Caruana}, \bibinfo{person}{Robert DeLine}, {and}
  \bibinfo{person}{Steven~M Drucker}.} \bibinfo{year}{2019}\natexlab{}.
\newblock \showarticletitle{Gamut: A design probe to understand how data
  scientists understand machine learning models}. In
  \bibinfo{booktitle}{\emph{Proceedings of the 2019 CHI Conference on Human
  Factors in Computing Systems}}. \bibinfo{pages}{1--13}.
\newblock


\bibitem[\protect\citeauthoryear{Hong, Hullman, and Bertini}{Hong
  et~al\mbox{.}}{2020}]%
        {hong2020human}
\bibfield{author}{\bibinfo{person}{Sungsoo~Ray Hong}, \bibinfo{person}{Jessica
  Hullman}, {and} \bibinfo{person}{Enrico Bertini}.}
  \bibinfo{year}{2020}\natexlab{}.
\newblock \showarticletitle{Human factors in model interpretability: Industry
  practices, challenges, and needs}.
\newblock \bibinfo{journal}{\emph{Proceedings of the ACM on Human-Computer
  Interaction}} \bibinfo{volume}{4}, \bibinfo{number}{CSCW1}
  (\bibinfo{year}{2020}), \bibinfo{pages}{1--26}.
\newblock


\bibitem[\protect\citeauthoryear{Kahneman}{Kahneman}{2011}]%
        {kahneman2011thinking}
\bibfield{author}{\bibinfo{person}{Daniel Kahneman}.}
  \bibinfo{year}{2011}\natexlab{}.
\newblock \bibinfo{booktitle}{\emph{Thinking, fast and slow}}.
\newblock \bibinfo{publisher}{Macmillan}.
\newblock


\bibitem[\protect\citeauthoryear{Kaur, Nori, Jenkins, Caruana, Wallach, and
  Wortman~Vaughan}{Kaur et~al\mbox{.}}{2020}]%
        {kaur2020interpreting}
\bibfield{author}{\bibinfo{person}{Harmanpreet Kaur}, \bibinfo{person}{Harsha
  Nori}, \bibinfo{person}{Samuel Jenkins}, \bibinfo{person}{Rich Caruana},
  \bibinfo{person}{Hanna Wallach}, {and} \bibinfo{person}{Jennifer
  Wortman~Vaughan}.} \bibinfo{year}{2020}\natexlab{}.
\newblock \showarticletitle{Interpreting Interpretability: Understanding Data
  Scientists' Use of Interpretability Tools for Machine Learning}. In
  \bibinfo{booktitle}{\emph{Proceedings of the 2020 CHI Conference on Human
  Factors in Computing Systems}}. \bibinfo{pages}{1--14}.
\newblock


\bibitem[\protect\citeauthoryear{Kim, Khanna, and Koyejo}{Kim
  et~al\mbox{.}}{2016}]%
        {kim2016examples}
\bibfield{author}{\bibinfo{person}{Been Kim}, \bibinfo{person}{Rajiv Khanna},
  {and} \bibinfo{person}{Oluwasanmi~O Koyejo}.}
  \bibinfo{year}{2016}\natexlab{}.
\newblock \showarticletitle{Examples are not enough, learn to criticize!
  criticism for interpretability}. In \bibinfo{booktitle}{\emph{Proceedings of
  NIPS}}.
\newblock


\bibitem[\protect\citeauthoryear{Kim, Wattenberg, Gilmer, Cai, Wexler, Viegas,
  et~al\mbox{.}}{Kim et~al\mbox{.}}{2018}]%
        {kim2018interpretability}
\bibfield{author}{\bibinfo{person}{Been Kim}, \bibinfo{person}{Martin
  Wattenberg}, \bibinfo{person}{Justin Gilmer}, \bibinfo{person}{Carrie Cai},
  \bibinfo{person}{James Wexler}, \bibinfo{person}{Fernanda Viegas},
  {et~al\mbox{.}}} \bibinfo{year}{2018}\natexlab{}.
\newblock \showarticletitle{Interpretability beyond feature attribution:
  Quantitative testing with concept activation vectors (tcav)}. In
  \bibinfo{booktitle}{\emph{International conference on machine learning}}.
  PMLR, \bibinfo{pages}{2668--2677}.
\newblock


\bibitem[\protect\citeauthoryear{Krause, Perer, and Ng}{Krause
  et~al\mbox{.}}{2016}]%
        {krause2016interacting}
\bibfield{author}{\bibinfo{person}{Josua Krause}, \bibinfo{person}{Adam Perer},
  {and} \bibinfo{person}{Kenney Ng}.} \bibinfo{year}{2016}\natexlab{}.
\newblock \showarticletitle{Interacting with predictions: Visual inspection of
  black-box machine learning models}. In \bibinfo{booktitle}{\emph{Proceedings
  of the 2016 CHI Conference on Human Factors in Computing Systems}}. ACM,
  \bibinfo{pages}{5686--5697}.
\newblock


\bibitem[\protect\citeauthoryear{Lakkaraju, Bach, and Leskovec}{Lakkaraju
  et~al\mbox{.}}{2016}]%
        {lakkaraju2016interpretable}
\bibfield{author}{\bibinfo{person}{Himabindu Lakkaraju},
  \bibinfo{person}{Stephen~H Bach}, {and} \bibinfo{person}{Jure Leskovec}.}
  \bibinfo{year}{2016}\natexlab{}.
\newblock \showarticletitle{Interpretable decision sets: A joint framework for
  description and prediction}. In \bibinfo{booktitle}{\emph{Proceedings of the
  22nd ACM SIGKDD international conference on knowledge discovery and data
  mining}}. \bibinfo{pages}{1675--1684}.
\newblock


\bibitem[\protect\citeauthoryear{Lakkaraju, Kamar, Caruana, and
  Leskovec}{Lakkaraju et~al\mbox{.}}{2017}]%
        {lakkaraju2017interpretable}
\bibfield{author}{\bibinfo{person}{Himabindu Lakkaraju}, \bibinfo{person}{Ece
  Kamar}, \bibinfo{person}{Rich Caruana}, {and} \bibinfo{person}{Jure
  Leskovec}.} \bibinfo{year}{2017}\natexlab{}.
\newblock \showarticletitle{Interpretable \& explorable approximations of black
  box models}.
\newblock \bibinfo{journal}{\emph{arXiv preprint arXiv:1707.01154}}
  (\bibinfo{year}{2017}).
\newblock


\bibitem[\protect\citeauthoryear{Leary and Kowalski}{Leary and
  Kowalski}{1990}]%
        {leary1990impression}
\bibfield{author}{\bibinfo{person}{Mark~R Leary} {and} \bibinfo{person}{Robin~M
  Kowalski}.} \bibinfo{year}{1990}\natexlab{}.
\newblock \showarticletitle{Impression management: A literature review and
  two-component model.}
\newblock \bibinfo{journal}{\emph{Psychological bulletin}}
  \bibinfo{volume}{107}, \bibinfo{number}{1} (\bibinfo{year}{1990}),
  \bibinfo{pages}{34}.
\newblock


\bibitem[\protect\citeauthoryear{Lei, G’Sell, Rinaldo, Tibshirani, and
  Wasserman}{Lei et~al\mbox{.}}{2018}]%
        {lei2018distribution}
\bibfield{author}{\bibinfo{person}{Jing Lei}, \bibinfo{person}{Max G’Sell},
  \bibinfo{person}{Alessandro Rinaldo}, \bibinfo{person}{Ryan~J Tibshirani},
  {and} \bibinfo{person}{Larry Wasserman}.} \bibinfo{year}{2018}\natexlab{}.
\newblock \showarticletitle{Distribution-free predictive inference for
  regression}.
\newblock \bibinfo{journal}{\emph{J. Amer. Statist. Assoc.}}
  \bibinfo{volume}{113}, \bibinfo{number}{523} (\bibinfo{year}{2018}),
  \bibinfo{pages}{1094--1111}.
\newblock


\bibitem[\protect\citeauthoryear{Li, Monroe, and Jurafsky}{Li
  et~al\mbox{.}}{2016}]%
        {li2016understanding}
\bibfield{author}{\bibinfo{person}{Jiwei Li}, \bibinfo{person}{Will Monroe},
  {and} \bibinfo{person}{Dan Jurafsky}.} \bibinfo{year}{2016}\natexlab{}.
\newblock \showarticletitle{Understanding neural networks through
  representation erasure}.
\newblock \bibinfo{journal}{\emph{arXiv preprint arXiv:1612.08220}}
  (\bibinfo{year}{2016}).
\newblock


\bibitem[\protect\citeauthoryear{Liao, Gruen, and Miller}{Liao
  et~al\mbox{.}}{2020}]%
        {liao2020questioning}
\bibfield{author}{\bibinfo{person}{Q~Vera Liao}, \bibinfo{person}{Daniel
  Gruen}, {and} \bibinfo{person}{Sarah Miller}.}
  \bibinfo{year}{2020}\natexlab{}.
\newblock \showarticletitle{Questioning the AI: informing design practices for
  explainable AI user experiences}. In \bibinfo{booktitle}{\emph{Proceedings of
  the 2020 CHI Conference on Human Factors in Computing Systems}}.
  \bibinfo{pages}{1--15}.
\newblock


\bibitem[\protect\citeauthoryear{Liao, Pribi{\'c}, Han, Miller, and Sow}{Liao
  et~al\mbox{.}}{2021}]%
        {liao2021question}
\bibfield{author}{\bibinfo{person}{Q~Vera Liao}, \bibinfo{person}{Milena
  Pribi{\'c}}, \bibinfo{person}{Jaesik Han}, \bibinfo{person}{Sarah Miller},
  {and} \bibinfo{person}{Daby Sow}.} \bibinfo{year}{2021}\natexlab{}.
\newblock \showarticletitle{Question-Driven Design Process for Explainable AI
  User Experiences}.
\newblock \bibinfo{journal}{\emph{arXiv preprint arXiv:2104.03483}}
  (\bibinfo{year}{2021}).
\newblock


\bibitem[\protect\citeauthoryear{Lim and Dey}{Lim and Dey}{2009}]%
        {lim2009assessing}
\bibfield{author}{\bibinfo{person}{Brian~Y Lim} {and} \bibinfo{person}{Anind~K
  Dey}.} \bibinfo{year}{2009}\natexlab{}.
\newblock \showarticletitle{Assessing demand for intelligibility in
  context-aware applications}. In \bibinfo{booktitle}{\emph{Proceedings of the
  11th international conference on Ubiquitous computing}}.
  \bibinfo{pages}{195--204}.
\newblock


\bibitem[\protect\citeauthoryear{Lipton}{Lipton}{2018}]%
        {lipton2018mythos}
\bibfield{author}{\bibinfo{person}{Zachary~C Lipton}.}
  \bibinfo{year}{2018}\natexlab{}.
\newblock \showarticletitle{The mythos of model interpretability}.
\newblock \bibinfo{journal}{\emph{Queue}} \bibinfo{volume}{16},
  \bibinfo{number}{3} (\bibinfo{year}{2018}), \bibinfo{pages}{31--57}.
\newblock


\bibitem[\protect\citeauthoryear{Looveren and Klaise}{Looveren and
  Klaise}{2021}]%
        {looveren2021interpretable}
\bibfield{author}{\bibinfo{person}{Arnaud~Van Looveren} {and}
  \bibinfo{person}{Janis Klaise}.} \bibinfo{year}{2021}\natexlab{}.
\newblock \showarticletitle{Interpretable counterfactual explanations guided by
  prototypes}. In \bibinfo{booktitle}{\emph{Joint European Conference on
  Machine Learning and Knowledge Discovery in Databases}}. Springer,
  \bibinfo{pages}{650--665}.
\newblock


\bibitem[\protect\citeauthoryear{Lucic, Haned, and de~Rijke}{Lucic
  et~al\mbox{.}}{2020}]%
        {lucic2020does}
\bibfield{author}{\bibinfo{person}{Ana Lucic}, \bibinfo{person}{Hinda Haned},
  {and} \bibinfo{person}{Maarten de Rijke}.} \bibinfo{year}{2020}\natexlab{}.
\newblock \showarticletitle{Why does my model fail? contrastive local
  explanations for retail forecasting}. In
  \bibinfo{booktitle}{\emph{Proceedings of the 2020 Conference on Fairness,
  Accountability, and Transparency}}. \bibinfo{pages}{90--98}.
\newblock


\bibitem[\protect\citeauthoryear{Lundberg, Erion, Chen, DeGrave, Prutkin, Nair,
  Katz, Himmelfarb, Bansal, and Lee}{Lundberg et~al\mbox{.}}{2020}]%
        {lundberg2020local}
\bibfield{author}{\bibinfo{person}{Scott~M Lundberg}, \bibinfo{person}{Gabriel
  Erion}, \bibinfo{person}{Hugh Chen}, \bibinfo{person}{Alex DeGrave},
  \bibinfo{person}{Jordan~M Prutkin}, \bibinfo{person}{Bala Nair},
  \bibinfo{person}{Ronit Katz}, \bibinfo{person}{Jonathan Himmelfarb},
  \bibinfo{person}{Nisha Bansal}, {and} \bibinfo{person}{Su-In Lee}.}
  \bibinfo{year}{2020}\natexlab{}.
\newblock \showarticletitle{From local explanations to global understanding
  with explainable AI for trees}.
\newblock \bibinfo{journal}{\emph{Nature machine intelligence}}
  \bibinfo{volume}{2}, \bibinfo{number}{1} (\bibinfo{year}{2020}),
  \bibinfo{pages}{56--67}.
\newblock


\bibitem[\protect\citeauthoryear{Lundberg and Lee}{Lundberg and Lee}{2017}]%
        {lundberg2017unified}
\bibfield{author}{\bibinfo{person}{Scott~M Lundberg} {and}
  \bibinfo{person}{Su-In Lee}.} \bibinfo{year}{2017}\natexlab{}.
\newblock \showarticletitle{A unified approach to interpreting model
  predictions}. In \bibinfo{booktitle}{\emph{Proceedings of the 31st
  international conference on neural information processing systems}}.
  \bibinfo{pages}{4768--4777}.
\newblock


\bibitem[\protect\citeauthoryear{Madumal, Miller, Sonenberg, and
  Vetere}{Madumal et~al\mbox{.}}{2019}]%
        {madumal2019grounded}
\bibfield{author}{\bibinfo{person}{Prashan Madumal}, \bibinfo{person}{Tim
  Miller}, \bibinfo{person}{Liz Sonenberg}, {and} \bibinfo{person}{Frank
  Vetere}.} \bibinfo{year}{2019}\natexlab{}.
\newblock \showarticletitle{A grounded interaction protocol for explainable
  artificial intelligence}.
\newblock \bibinfo{journal}{\emph{arXiv preprint arXiv:1903.02409}}
  (\bibinfo{year}{2019}).
\newblock


\bibitem[\protect\citeauthoryear{Malle}{Malle}{2006}]%
        {malle2006mind}
\bibfield{author}{\bibinfo{person}{Bertram~F Malle}.}
  \bibinfo{year}{2006}\natexlab{}.
\newblock \bibinfo{booktitle}{\emph{How the mind explains behavior: Folk
  explanations, meaning, and social interaction}}.
\newblock \bibinfo{publisher}{Mit Press}.
\newblock


\bibitem[\protect\citeauthoryear{Miller}{Miller}{2018}]%
        {miller2018explanation}
\bibfield{author}{\bibinfo{person}{Tim Miller}.}
  \bibinfo{year}{2018}\natexlab{}.
\newblock \showarticletitle{Explanation in artificial intelligence: Insights
  from the social sciences}.
\newblock \bibinfo{journal}{\emph{Artificial Intelligence}}
  (\bibinfo{year}{2018}).
\newblock


\bibitem[\protect\citeauthoryear{Miller, Howe, and Sonenberg}{Miller
  et~al\mbox{.}}{2017}]%
        {miller2017explainable}
\bibfield{author}{\bibinfo{person}{Tim Miller}, \bibinfo{person}{Piers Howe},
  {and} \bibinfo{person}{Liz Sonenberg}.} \bibinfo{year}{2017}\natexlab{}.
\newblock \showarticletitle{Explainable AI: Beware of inmates running the
  asylum or: How I learnt to stop worrying and love the social and behavioural
  sciences}.
\newblock \bibinfo{journal}{\emph{arXiv preprint arXiv:1712.00547}}
  (\bibinfo{year}{2017}).
\newblock


\bibitem[\protect\citeauthoryear{Mitchell, Wu, Zaldivar, Barnes, Vasserman,
  Hutchinson, Spitzer, Raji, and Gebru}{Mitchell et~al\mbox{.}}{2019}]%
        {mitchell2019model}
\bibfield{author}{\bibinfo{person}{Margaret Mitchell}, \bibinfo{person}{Simone
  Wu}, \bibinfo{person}{Andrew Zaldivar}, \bibinfo{person}{Parker Barnes},
  \bibinfo{person}{Lucy Vasserman}, \bibinfo{person}{Ben Hutchinson},
  \bibinfo{person}{Elena Spitzer}, \bibinfo{person}{Inioluwa~Deborah Raji},
  {and} \bibinfo{person}{Timnit Gebru}.} \bibinfo{year}{2019}\natexlab{}.
\newblock \showarticletitle{Model cards for model reporting}. In
  \bibinfo{booktitle}{\emph{Proceedings of the conference on fairness,
  accountability, and transparency}}. \bibinfo{pages}{220--229}.
\newblock


\bibitem[\protect\citeauthoryear{Mothilal, Sharma, and Tan}{Mothilal
  et~al\mbox{.}}{2019}]%
        {mothilal2019explaining}
\bibfield{author}{\bibinfo{person}{Ramaravind~Kommiya Mothilal},
  \bibinfo{person}{Amit Sharma}, {and} \bibinfo{person}{Chenhao Tan}.}
  \bibinfo{year}{2019}\natexlab{}.
\newblock \showarticletitle{Explaining machine learning classifiers through
  diverse counterfactual explanations}.
\newblock \bibinfo{journal}{\emph{arXiv preprint arXiv:1905.07697}}
  (\bibinfo{year}{2019}).
\newblock


\bibitem[\protect\citeauthoryear{Narkar, Zhang, Liao, Wang, and Weisz}{Narkar
  et~al\mbox{.}}{2021}]%
        {narkar2021model}
\bibfield{author}{\bibinfo{person}{Shweta Narkar}, \bibinfo{person}{Yunfeng
  Zhang}, \bibinfo{person}{Q~Vera Liao}, \bibinfo{person}{Dakuo Wang}, {and}
  \bibinfo{person}{Justin~D Weisz}.} \bibinfo{year}{2021}\natexlab{}.
\newblock \showarticletitle{Model LineUpper: Supporting Interactive Model
  Comparison at Multiple Levels for AutoML}. In \bibinfo{booktitle}{\emph{26th
  International Conference on Intelligent User Interfaces}}.
  \bibinfo{pages}{170--174}.
\newblock


\bibitem[\protect\citeauthoryear{Norman}{Norman}{2013}]%
        {norman2013design}
\bibfield{author}{\bibinfo{person}{Don Norman}.}
  \bibinfo{year}{2013}\natexlab{}.
\newblock \bibinfo{booktitle}{\emph{The design of everyday things: Revised and
  expanded edition}}.
\newblock \bibinfo{publisher}{Basic books}.
\newblock


\bibitem[\protect\citeauthoryear{Nourani, Roy, Block, Honeycutt, Rahman, Ragan,
  and Gogate}{Nourani et~al\mbox{.}}{2021}]%
        {nourani2021anchoring}
\bibfield{author}{\bibinfo{person}{Mahsan Nourani}, \bibinfo{person}{Chiradeep
  Roy}, \bibinfo{person}{Jeremy~E Block}, \bibinfo{person}{Donald~R Honeycutt},
  \bibinfo{person}{Tahrima Rahman}, \bibinfo{person}{Eric Ragan}, {and}
  \bibinfo{person}{Vibhav Gogate}.} \bibinfo{year}{2021}\natexlab{}.
\newblock \showarticletitle{Anchoring Bias Affects Mental Model Formation and
  User Reliance in Explainable AI Systems}. In \bibinfo{booktitle}{\emph{26th
  International Conference on Intelligent User Interfaces}}.
  \bibinfo{pages}{340--350}.
\newblock


\bibitem[\protect\citeauthoryear{P{\'a}ez}{P{\'a}ez}{2019}]%
        {paez2019pragmatic}
\bibfield{author}{\bibinfo{person}{Andr{\'e}s P{\'a}ez}.}
  \bibinfo{year}{2019}\natexlab{}.
\newblock \showarticletitle{The pragmatic turn in explainable artificial
  intelligence (XAI)}.
\newblock \bibinfo{journal}{\emph{Minds and Machines}} \bibinfo{volume}{29},
  \bibinfo{number}{3} (\bibinfo{year}{2019}), \bibinfo{pages}{441--459}.
\newblock


\bibitem[\protect\citeauthoryear{Papernot and McDaniel}{Papernot and
  McDaniel}{2018}]%
        {papernot2018deep}
\bibfield{author}{\bibinfo{person}{Nicolas Papernot} {and}
  \bibinfo{person}{Patrick McDaniel}.} \bibinfo{year}{2018}\natexlab{}.
\newblock \showarticletitle{Deep k-nearest neighbors: Towards confident,
  interpretable and robust deep learning}.
\newblock \bibinfo{journal}{\emph{arXiv preprint arXiv:1803.04765}}
  (\bibinfo{year}{2018}).
\newblock


\bibitem[\protect\citeauthoryear{Petty and Cacioppo}{Petty and
  Cacioppo}{1986}]%
        {petty1986elaboration}
\bibfield{author}{\bibinfo{person}{Richard~E Petty} {and}
  \bibinfo{person}{John~T Cacioppo}.} \bibinfo{year}{1986}\natexlab{}.
\newblock \showarticletitle{The elaboration likelihood model of persuasion}.
\newblock In \bibinfo{booktitle}{\emph{Communication and persuasion}}.
  \bibinfo{publisher}{Springer}, \bibinfo{pages}{1--24}.
\newblock


\bibitem[\protect\citeauthoryear{Poursabzi-Sangdeh, Goldstein, Hofman,
  Wortman~Vaughan, and Wallach}{Poursabzi-Sangdeh et~al\mbox{.}}{2021}]%
        {poursabzi2021manipulating}
\bibfield{author}{\bibinfo{person}{Forough Poursabzi-Sangdeh},
  \bibinfo{person}{Daniel~G Goldstein}, \bibinfo{person}{Jake~M Hofman},
  \bibinfo{person}{Jennifer~Wortman Wortman~Vaughan}, {and}
  \bibinfo{person}{Hanna Wallach}.} \bibinfo{year}{2021}\natexlab{}.
\newblock \showarticletitle{Manipulating and measuring model interpretability}.
  In \bibinfo{booktitle}{\emph{Proceedings of the 2021 CHI Conference on Human
  Factors in Computing Systems}}. \bibinfo{pages}{1--52}.
\newblock


\bibitem[\protect\citeauthoryear{Preece, Harborne, Braines, Tomsett, and
  Chakraborty}{Preece et~al\mbox{.}}{2018}]%
        {preece2018stakeholders}
\bibfield{author}{\bibinfo{person}{Alun Preece}, \bibinfo{person}{Dan
  Harborne}, \bibinfo{person}{Dave Braines}, \bibinfo{person}{Richard Tomsett},
  {and} \bibinfo{person}{Supriyo Chakraborty}.}
  \bibinfo{year}{2018}\natexlab{}.
\newblock \showarticletitle{Stakeholders in explainable AI}.
\newblock \bibinfo{journal}{\emph{arXiv preprint arXiv:1810.00184}}
  (\bibinfo{year}{2018}).
\newblock


\bibitem[\protect\citeauthoryear{Puri, Dhurandhar, Pedapati, Shanmugam, Wei,
  and Varshney}{Puri et~al\mbox{.}}{2021}]%
        {puri2021cofrnets}
\bibfield{author}{\bibinfo{person}{Isha Puri}, \bibinfo{person}{Amit
  Dhurandhar}, \bibinfo{person}{Tejaswini Pedapati},
  \bibinfo{person}{Karthikeyan Shanmugam}, \bibinfo{person}{Dennis Wei}, {and}
  \bibinfo{person}{Kush~R. Varshney}.} \bibinfo{year}{2021}\natexlab{}.
\newblock \showarticletitle{CoFrNets: Interpretable neural architecture
  inspired by continued fractions}. In \bibinfo{booktitle}{\emph{Advances in
  Neural Information Processing Systems}}.
\newblock


\bibitem[\protect\citeauthoryear{Rastogi, Zhang, Wei, Varshney, Dhurandhar, and
  Tomsett}{Rastogi et~al\mbox{.}}{2022}]%
        {rastogi2022deciding}
\bibfield{author}{\bibinfo{person}{Charvi Rastogi}, \bibinfo{person}{Yunfeng
  Zhang}, \bibinfo{person}{Dennis Wei}, \bibinfo{person}{Kush~R. Varshney},
  \bibinfo{person}{Amit Dhurandhar}, {and} \bibinfo{person}{Richard Tomsett}.}
  \bibinfo{year}{2022}\natexlab{}.
\newblock \showarticletitle{Deciding Fast and Slow: The Role of Cognitive
  Biases in AI-Assisted Decision-Making}. In
  \bibinfo{booktitle}{\emph{Proceedings of the ACM Conference on Computer
  Supported Cooperative Work and Social Computing}}.
\newblock


\bibitem[\protect\citeauthoryear{Ribeiro, Singh, and Guestrin}{Ribeiro
  et~al\mbox{.}}{2016}]%
        {ribeiro2016should}
\bibfield{author}{\bibinfo{person}{Marco~Tulio Ribeiro},
  \bibinfo{person}{Sameer Singh}, {and} \bibinfo{person}{Carlos Guestrin}.}
  \bibinfo{year}{2016}\natexlab{}.
\newblock \showarticletitle{Why should i trust you?: Explaining the predictions
  of any classifier}. In \bibinfo{booktitle}{\emph{Proceedings of KDD}}.
\newblock


\bibitem[\protect\citeauthoryear{Ribeiro, Singh, and Guestrin}{Ribeiro
  et~al\mbox{.}}{2018}]%
        {ribeiro2018anchors}
\bibfield{author}{\bibinfo{person}{Marco~Tulio Ribeiro},
  \bibinfo{person}{Sameer Singh}, {and} \bibinfo{person}{Carlos Guestrin}.}
  \bibinfo{year}{2018}\natexlab{}.
\newblock \showarticletitle{Anchors: High-precision model-agnostic
  explanations}. In \bibinfo{booktitle}{\emph{Thirty-Second AAAI Conference on
  Artificial Intelligence}}.
\newblock


\bibitem[\protect\citeauthoryear{Rieh and Danielson}{Rieh and
  Danielson}{2007}]%
        {rieh2007credibility}
\bibfield{author}{\bibinfo{person}{Soo~Young Rieh} {and}
  \bibinfo{person}{David~R Danielson}.} \bibinfo{year}{2007}\natexlab{}.
\newblock \showarticletitle{Credibility: A multidisciplinary framework}.
\newblock \bibinfo{journal}{\emph{Annual review of information science and
  technology}} \bibinfo{volume}{41}, \bibinfo{number}{1}
  (\bibinfo{year}{2007}), \bibinfo{pages}{307--364}.
\newblock


\bibitem[\protect\citeauthoryear{Robertson, Kokkinakis, Hook, Kirman, Block,
  Ursu, Patra, Demediuk, Drachen, and Olarewaju}{Robertson
  et~al\mbox{.}}{2021}]%
        {robertson2021wait}
\bibfield{author}{\bibinfo{person}{Justus Robertson},
  \bibinfo{person}{Athanasios~Vasileios Kokkinakis}, \bibinfo{person}{Jonathan
  Hook}, \bibinfo{person}{Ben Kirman}, \bibinfo{person}{Florian Block},
  \bibinfo{person}{Marian~F Ursu}, \bibinfo{person}{Sagarika Patra},
  \bibinfo{person}{Simon Demediuk}, \bibinfo{person}{Anders Drachen}, {and}
  \bibinfo{person}{Oluseyi Olarewaju}.} \bibinfo{year}{2021}\natexlab{}.
\newblock \showarticletitle{Wait, But Why?: Assessing Behavior Explanation
  Strategies for Real-Time Strategy Games}. In \bibinfo{booktitle}{\emph{26th
  International Conference on Intelligent User Interfaces}}.
  \bibinfo{pages}{32--42}.
\newblock


\bibitem[\protect\citeauthoryear{Rudin}{Rudin}{2019}]%
        {rudin2019stop}
\bibfield{author}{\bibinfo{person}{Cynthia Rudin}.}
  \bibinfo{year}{2019}\natexlab{}.
\newblock \showarticletitle{Stop explaining black box machine learning models
  for high stakes decisions and use interpretable models instead}.
\newblock \bibinfo{journal}{\emph{Nature Machine Intelligence}}
  \bibinfo{volume}{1}, \bibinfo{number}{5} (\bibinfo{year}{2019}),
  \bibinfo{pages}{206--215}.
\newblock


\bibitem[\protect\citeauthoryear{Selvaraju, Cogswell, Das, Vedantam, Parikh,
  and Batra}{Selvaraju et~al\mbox{.}}{2017}]%
        {selvaraju2017grad}
\bibfield{author}{\bibinfo{person}{Ramprasaath~R Selvaraju},
  \bibinfo{person}{Michael Cogswell}, \bibinfo{person}{Abhishek Das},
  \bibinfo{person}{Ramakrishna Vedantam}, \bibinfo{person}{Devi Parikh}, {and}
  \bibinfo{person}{Dhruv Batra}.} \bibinfo{year}{2017}\natexlab{}.
\newblock \showarticletitle{Grad-cam: Visual explanations from deep networks
  via gradient-based localization}. In \bibinfo{booktitle}{\emph{Proceedings of
  the IEEE international conference on computer vision}}.
  \bibinfo{pages}{618--626}.
\newblock


\bibitem[\protect\citeauthoryear{Springer and Whittaker}{Springer and
  Whittaker}{2019}]%
        {springer2019progressive}
\bibfield{author}{\bibinfo{person}{Aaron Springer} {and} \bibinfo{person}{Steve
  Whittaker}.} \bibinfo{year}{2019}\natexlab{}.
\newblock \showarticletitle{Progressive disclosure: empirically motivated
  approaches to designing effective transparency}. In
  \bibinfo{booktitle}{\emph{Proceedings of the 24th international conference on
  intelligent user interfaces}}. \bibinfo{pages}{107--120}.
\newblock


\bibitem[\protect\citeauthoryear{Suresh, Gomez, Nam, and Satyanarayan}{Suresh
  et~al\mbox{.}}{2021}]%
        {suresh2021beyond}
\bibfield{author}{\bibinfo{person}{Harini Suresh}, \bibinfo{person}{Steven~R
  Gomez}, \bibinfo{person}{Kevin~K Nam}, {and} \bibinfo{person}{Arvind
  Satyanarayan}.} \bibinfo{year}{2021}\natexlab{}.
\newblock \showarticletitle{Beyond Expertise and Roles: A Framework to
  Characterize the Stakeholders of Interpretable Machine Learning and their
  Needs}. In \bibinfo{booktitle}{\emph{Proceedings of the 2021 CHI Conference
  on Human Factors in Computing Systems}}. \bibinfo{pages}{1--16}.
\newblock


\bibitem[\protect\citeauthoryear{Szymanski, Millecamp, and Verbert}{Szymanski
  et~al\mbox{.}}{2021}]%
        {szymanski2021visual}
\bibfield{author}{\bibinfo{person}{Maxwell Szymanski}, \bibinfo{person}{Martijn
  Millecamp}, {and} \bibinfo{person}{Katrien Verbert}.}
  \bibinfo{year}{2021}\natexlab{}.
\newblock \showarticletitle{Visual, textual or hybrid: the effect of user
  expertise on different explanations}. In \bibinfo{booktitle}{\emph{26th
  International Conference on Intelligent User Interfaces}}.
  \bibinfo{pages}{109--119}.
\newblock


\bibitem[\protect\citeauthoryear{Tan, Caruana, Hooker, Koch, and Gordo}{Tan
  et~al\mbox{.}}{2018}]%
        {tan2018learning}
\bibfield{author}{\bibinfo{person}{Sarah Tan}, \bibinfo{person}{Rich Caruana},
  \bibinfo{person}{Giles Hooker}, \bibinfo{person}{Paul Koch}, {and}
  \bibinfo{person}{Albert Gordo}.} \bibinfo{year}{2018}\natexlab{}.
\newblock \showarticletitle{Learning global additive explanations for neural
  nets using model distillation}.
\newblock \bibinfo{journal}{\emph{arXiv preprint arXiv:1801.08640}}
  (\bibinfo{year}{2018}).
\newblock


\bibitem[\protect\citeauthoryear{Vaughan and Wallach}{Vaughan and
  Wallach}{2020}]%
        {vaughan2020human}
\bibfield{author}{\bibinfo{person}{Jennifer~Wortman Vaughan} {and}
  \bibinfo{person}{Hanna Wallach}.} \bibinfo{year}{2020}\natexlab{}.
\newblock \showarticletitle{A human-centered agenda for intelligible machine
  learning}.
\newblock \bibinfo{journal}{\emph{Machines We Trust: Getting Along with
  Artificial Intelligence}} (\bibinfo{year}{2020}).
\newblock


\bibitem[\protect\citeauthoryear{Wachter, Mittelstadt, and Russell}{Wachter
  et~al\mbox{.}}{2017}]%
        {wachter2017counterfactual}
\bibfield{author}{\bibinfo{person}{Sandra Wachter}, \bibinfo{person}{Brent
  Mittelstadt}, {and} \bibinfo{person}{Chris Russell}.}
  \bibinfo{year}{2017}\natexlab{}.
\newblock \showarticletitle{Counterfactual explanations without opening the
  black box: Automated decisions and the GDPR}.
\newblock


\bibitem[\protect\citeauthoryear{Walton}{Walton}{2004}]%
        {walton2004new}
\bibfield{author}{\bibinfo{person}{Douglas Walton}.}
  \bibinfo{year}{2004}\natexlab{}.
\newblock \showarticletitle{A new dialectical theory of explanation}.
\newblock \bibinfo{journal}{\emph{Philosophical Explorations}}
  \bibinfo{volume}{7}, \bibinfo{number}{1} (\bibinfo{year}{2004}),
  \bibinfo{pages}{71--89}.
\newblock


\bibitem[\protect\citeauthoryear{Wang, Yang, Abdul, and Lim}{Wang
  et~al\mbox{.}}{2019}]%
        {wang2019designing}
\bibfield{author}{\bibinfo{person}{Danding Wang}, \bibinfo{person}{Qian Yang},
  \bibinfo{person}{Ashraf Abdul}, {and} \bibinfo{person}{Brian~Y Lim}.}
  \bibinfo{year}{2019}\natexlab{}.
\newblock \showarticletitle{Designing Theory-Driven User-Centric Explainable
  AI}. In \bibinfo{booktitle}{\emph{Proceedings of the 2019 CHI Conference on
  Human Factors in Computing Systems}}. ACM, \bibinfo{pages}{601}.
\newblock


\bibitem[\protect\citeauthoryear{Wang and Yin}{Wang and Yin}{2021}]%
        {wang2021explanations}
\bibfield{author}{\bibinfo{person}{Xinru Wang} {and} \bibinfo{person}{Ming
  Yin}.} \bibinfo{year}{2021}\natexlab{}.
\newblock \showarticletitle{Are Explanations Helpful? A Comparative Study of
  the Effects of Explanations in AI-Assisted Decision-Making}. In
  \bibinfo{booktitle}{\emph{26th International Conference on Intelligent User
  Interfaces}}. \bibinfo{pages}{318--328}.
\newblock


\bibitem[\protect\citeauthoryear{Wei, Dash, Gao, and Gunluk}{Wei
  et~al\mbox{.}}{2019}]%
        {wei2019generalized}
\bibfield{author}{\bibinfo{person}{Dennis Wei}, \bibinfo{person}{Sanjeeb Dash},
  \bibinfo{person}{Tian Gao}, {and} \bibinfo{person}{Oktay Gunluk}.}
  \bibinfo{year}{2019}\natexlab{}.
\newblock \showarticletitle{Generalized linear rule models}. In
  \bibinfo{booktitle}{\emph{International Conference on Machine Learning}}.
  PMLR, \bibinfo{pages}{6687--6696}.
\newblock


\bibitem[\protect\citeauthoryear{Wei, Lu, and Song}{Wei et~al\mbox{.}}{2015}]%
        {wei2015variable}
\bibfield{author}{\bibinfo{person}{Pengfei Wei}, \bibinfo{person}{Zhenzhou Lu},
  {and} \bibinfo{person}{Jingwen Song}.} \bibinfo{year}{2015}\natexlab{}.
\newblock \showarticletitle{Variable importance analysis: a comprehensive
  review}.
\newblock \bibinfo{journal}{\emph{Reliability Engineering \& System Safety}}
  \bibinfo{volume}{142} (\bibinfo{year}{2015}), \bibinfo{pages}{399--432}.
\newblock


\bibitem[\protect\citeauthoryear{Wilson}{Wilson}{1981}]%
        {wilson1981user}
\bibfield{author}{\bibinfo{person}{Tom~D Wilson}.}
  \bibinfo{year}{1981}\natexlab{}.
\newblock \showarticletitle{On user studies and information needs}.
\newblock \bibinfo{journal}{\emph{Journal of documentation}}
  (\bibinfo{year}{1981}).
\newblock


\bibitem[\protect\citeauthoryear{Xie, Chen, Kao, Gao, and Chen}{Xie
  et~al\mbox{.}}{2020}]%
        {xie2020chexplain}
\bibfield{author}{\bibinfo{person}{Yao Xie}, \bibinfo{person}{Melody Chen},
  \bibinfo{person}{David Kao}, \bibinfo{person}{Ge Gao}, {and}
  \bibinfo{person}{Xiang~`Anthony' Chen}.} \bibinfo{year}{2020}\natexlab{}.
\newblock \showarticletitle{CheXplain: Enabling Physicians to Explore and
  Understand Data-Driven, AI-Enabled Medical Imaging Analysis}. In
  \bibinfo{booktitle}{\emph{Proceedings of the 2020 CHI Conference on Human
  Factors in Computing Systems}}. \bibinfo{pages}{1--13}.
\newblock


\bibitem[\protect\citeauthoryear{Yeh, Kim, Yen, and Ravikumar}{Yeh
  et~al\mbox{.}}{2018}]%
        {yeh2018representer}
\bibfield{author}{\bibinfo{person}{Chih-Kuan Yeh}, \bibinfo{person}{Joon Kim},
  \bibinfo{person}{Ian En-Hsu Yen}, {and} \bibinfo{person}{Pradeep~K
  Ravikumar}.} \bibinfo{year}{2018}\natexlab{}.
\newblock \showarticletitle{Representer point selection for explaining deep
  neural networks}.
\newblock \bibinfo{journal}{\emph{Advances in neural information processing
  systems}}  \bibinfo{volume}{31} (\bibinfo{year}{2018}).
\newblock


\bibitem[\protect\citeauthoryear{Zhang, Liao, and Bellamy}{Zhang
  et~al\mbox{.}}{2020}]%
        {zhang2020effect}
\bibfield{author}{\bibinfo{person}{Yunfeng Zhang}, \bibinfo{person}{Q~Vera
  Liao}, {and} \bibinfo{person}{Rachel~KE Bellamy}.}
  \bibinfo{year}{2020}\natexlab{}.
\newblock \showarticletitle{Effect of confidence and explanation on accuracy
  and trust calibration in AI-assisted decision making}. In
  \bibinfo{booktitle}{\emph{Proceedings of the 2020 Conference on Fairness,
  Accountability, and Transparency}}. \bibinfo{pages}{295--305}.
\newblock


\end{thebibliography}

\end{document}